\documentclass[final,5p,times,twocolumn]{elsarticle}
\makeatletter
\def\ps@pprintTitle{%
 \let\@oddhead\@empty
 \let\@evenhead\@empty
 \let\@evenfoot\@oddfoot}
\makeatother

\usepackage[pagebackref=true,breaklinks=true,colorlinks,bookmarks=false]{hyperref}
\usepackage{amsmath,amssymb} 
\usepackage{bm}

\usepackage{lipsum}
\usepackage{algorithm}
\usepackage{enumitem}

\usepackage[noend]{algpseudocode}
\usepackage{booktabs}
\usepackage{xspace}
\usepackage[normalem]{ulem} 
\usepackage{wrapfig}
\usepackage{sidecap}
\usepackage{multirow}
\usepackage{colortbl}

\makeatletter
\DeclareRobustCommand\onedot{\futurelet\@let@token\@onedot}
\def\@onedot{\ifx\@let@token.\else.\null\fi\xspace}

\def\eg{\emph{e.g}\onedot} 
\def\ie{\emph{i.e}\onedot} 
 
\def\etc{\emph{etc}\onedot}

\makeatother

\makeatletter
\newcommand*\bigcdot{\mathpalette\bigcdot@{.5}}
\newcommand*\bigcdot@[2]{\mathbin{\vcenter{\hbox{\scalebox{#2}{$\m@th#1\bullet$}}}}}
  \newcommand\tabcaption{\def\@captype{table}\caption}
\makeatother

\definecolor{lightgray}{rgb}{0.83, 0.83, 0.83}
\newcommand{\figref}[1]{Fig.~\ref{#1}}
\newcommand{\equref}[1]{Eq.~\eqref{#1}}
\newcommand{\secref}[1]{Sec.~\ref{#1}}
\newcommand{\tabref}[1]{Table~\ref{#1}}

\begin{document}

\begin{frontmatter}

\title{\textbf{MUNet: Motion Uncertainty-aware Semi-supervised Video Object Segmentation}}
\author{Jiadai Sun$^{1*}$,\ 
        Yuxin Mao$^{1*}$,\
        Yuchao Dai$^{1**}$,\ 
        Yiran Zhong$^{2}$,
        Jianyuan Wang$^{2}$
    \\
    $^1$School of Electronics and Information, Northwestern Polytechnical University, Xi'an, China \\
    $^2$SenseTime, Shanghai, China. }

\tnotetext[mytitlenote]{The first two authors contributed equally.} 
\tnotetext[mytitlenote]{Yuchao Dai (daiyuchao@nwpu.edu.cn) is the corresponding author.}

\begin{abstract}
The task of semi-supervised video object segmentation (VOS) has been greatly advanced and state-of-the-art performance has been made by dense matching-based methods. 
The recent methods leverage space-time memory (STM) networks and learn to retrieve relevant information from all available sources, where the past frames with object masks form an external memory and the current frame as the query is segmented using the mask information in the memory. However, when forming the memory and performing matching, these methods only exploit the appearance information while ignoring the motion information. 
In this paper, we advocate the return of the \emph{motion information} and propose a motion uncertainty-aware framework (MUNet) for semi-supervised VOS. 
First, we propose an implicit method to learn the spatial correspondences between neighboring frames, building upon a correlation cost volume. 
To handle the challenging cases of occlusion and textureless regions during constructing dense correspondences, we incorporate the uncertainty in dense matching and achieve motion uncertainty-aware feature representation.
Second, we introduce a motion-aware spatial attention module to effectively fuse the motion feature with the semantic feature. 
Comprehensive experiments on challenging benchmarks show that 
\textbf{\textit{using a small amount of data and combining it with powerful motion information can bring a significant performance boost}}. We achieve ${76.5\%}$ $\mathcal{J} \& \mathcal{F}$ only using DAVIS17 for training, which significantly outperforms the \textit{SOTA} methods under the low-data protocol. \textit{The code will be released.}
\end{abstract}


\end{frontmatter}


\section{Introduction}

Video object segmentation (VOS) aims at segmenting the foreground objects from a given video with motion, which is a classic task in computer vision with many applications, including surveillance, video compression, and motion understanding \etc.
In this paper, we focus on the most practical and widely studied setting, \ie, semi-supervised VOS \cite{liang_AFB_URR_NIPS_2020,yang_CFBI_ECCV_2020,Robinson_frtm-vos_CVPR_2020,oh_STM_ICCV_2019, yin2021agunet_PR}, whereas the scope is to segment target objects over video sequences only given the initial mask of the first frame as prior and visual guidance. 
This is a challenging problem because the target objects can be confused with similar instances of the same category, and their appearance might vary drastically over time due to scale change, pose changes, fast motion, truncation, blurry effects, occlusions \etc. Essentially, these challenges could not be addressed with image appearance information only.

Recently, various deep learning based VOS approaches have been proposed \cite{sun2020adaptive_PR,zhao2021real_PR}, which could be roughly categorized as propagation-based methods \cite{Perazzi_masktrack_CVPR_2017, oh_seoung_fastvos_RGMP_CVPR_2018,li_vsreid_cvprw_davis_2017} and feature matching based methods \cite{oh_STM_ICCV_2019, Li_STM-cycle_NeurIPS_2020, liang_AFB_URR_NIPS_2020, lu_episodicvos_eccv_2020, seong_kernelizedvos_eccv_2020}. 
The former generally formulate the task as object mask propagation, while the latter leverages memory networks to retrieve relevant information. 
Nowadays, the feature matching-based methods such as Space-Time Memory (STM) Networks \cite{oh_STM_ICCV_2019,liang_AFB_URR_NIPS_2020,Li_STM-cycle_NeurIPS_2020,liyu_fast_vos_ECCV_2020, lu_episodicvos_eccv_2020, seong_kernelizedvos_eccv_2020} have achieved state-of-the-art performance in VOS. 
The key to the success lies in introducing the feature matching of historical frames using non-local operations with a well-designed feature-memory-bank mechanism. 
They conduct matching using all previous frames with the corresponding object segmentation results through feature similarity query, and infer the object mask of the current frame. 
However, these methods heavily rely on the matching of object appearance between frames, while motion information, as the critical feature between video frames, tends to be ignored.
Prior to the success of these feature matching-based methods, explicit motion modeling in the form of dense optical flow have been exploited \cite{cheng_segflow_cvpr_2017, jang_online_vos_ctn_cvpr_2017, Perazzi_masktrack_CVPR_2017,Dave_seg_anything_moves_ICCV_2019, khoreva_lucid_dream_DAVIS_2017,li_vsreid_cvprw_davis_2017, tian2020joint_PR}, where the dense optical flow is pre-computed from optical flow estimation networks \cite{ilg_flownet2_cvpr_2017,hui_liteflownet_cvpr_2018,revaud_epicflow_cvpr_2015, hu_efficient_flow_cvpr_2016, kroeger_fast_flow_eccv_2016, wang_DICL_NIPS_2020, teed_raft_eccv_2020}.
However, the explicit use of optical flow not only requires additional large dataset (having a domain gap with VOS datasets) training for an optical flow network but also cannot capture the critical challenges in VOS, \ie, occlusion, textureless regions, fast motion, and blurry effects.

\begin{figure*}[!t]
\label{fig:vis_compare_result}
\centering
    \includegraphics[width=\linewidth]{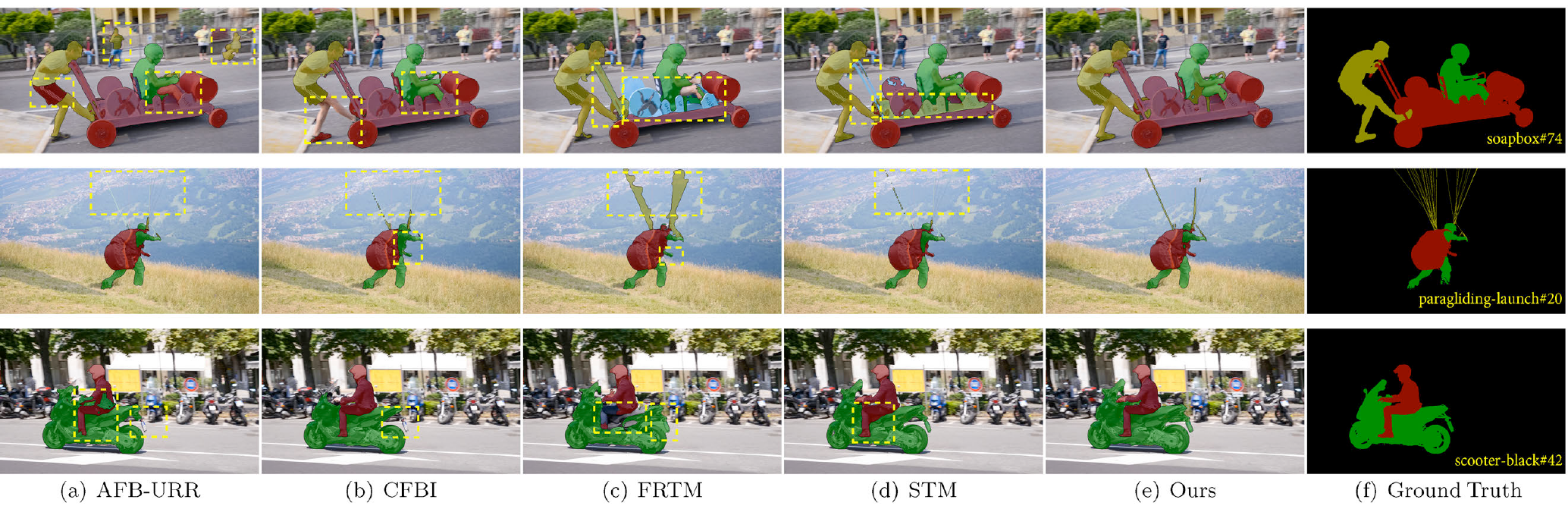}
    
    \caption{\textbf{Qualitative comparison} with competing methods \cite{liang_AFB_URR_NIPS_2020,yang_CFBI_ECCV_2020,Robinson_frtm-vos_CVPR_2020,oh_STM_ICCV_2019} on three sequences of DAVIS17 validation set. Through the implicit modeling of objects motion and uncertainty, we can obtain more accurate results, especially in multiple bodies movements (\textit{row\#1,3}) and thin lines (\textit{row\#2}). The inaccurate parts are marked with a yellow dashed box.}
    \vspace{-0.5\baselineskip}
\end{figure*}

In this paper, we advocate that VOS should not only rely on image appearance similarity matching but also emphasize the essential motion information from the video, which exists in any adjacent frames and will not disappear over time.
We propose a novel framework for semi-supervised VOS named \textbf{MUNet} that embeds motion information into a single branch pipeline without a pre-computed optical flow.
Given a video sequence, MUNet uses a dense matching based method with a feature-memory-bank to store both appearance and motion features, as shown in \figref{fig:OurFramework}.
To avoid problems caused by explicitly using optical flow, we propose a \textit{lightweight} \textbf{Motion Uncertainty-aware Layer (MU-Layer)} to \textit{implicitly} model motion information from adjacent frames. 
Specifically, we use a cost volume to model the displacement and motion uncertainty as a motion feature to establish spatio-temporal relationships, which is calculated from high-level semantic features of adjacent frames. 
In addition, we design a \textbf{Motion-aware Spatial Attention Module (MSAM)} to effectively fuse the appearance feature and the motion uncertainty-aware feature, then we use this module to guide the segmentation of video sequences. 
Different from the two-stream methods that need pre-computed optical flow as input,
MUNet \textit{only requires the raw images and does not need the supervision of optical flow}, which greatly expands the practicability and scope of application.
Meanwhile, under the experimental protocol using a small amount of data, the powerful motion information brings a significant performance boost without complex tricks.

We conduct comprehensive experiments on DAVIS17 and YouTube-VOS18.
Experimental results show that our method achieves state-of-the-art accuracy on the validation set of DAVIS17 ($\mathcal{J\&F} \ {76.5}\%$ ranks \textit{1st} under \texttt{protocol(1)}, ${81.1}\%$ ranks \textit{2nd} under \texttt{protocol(2)}, ${78.1}\%$ ranks \textit{1st} under \texttt{protocol(3-1)}), which exceeds all competing methods under the same settings.
We provide qualitative comparisons with four \textit{SOTA} methods \cite{liang_AFB_URR_NIPS_2020,yang_CFBI_ECCV_2020,Robinson_frtm-vos_CVPR_2020,oh_STM_ICCV_2019} in \figref{fig:vis_compare_result}. 
MUNet can accurately demarcate the boundaries of multiple objects and does not overly cover or ignore small structures.
As shown in \figref{fig:flow_uncertainty_shown}, the MU-Layer can implicitly learn a reasonable uncertainty map and displacement vector, and effectively activate the area of the moving object through the motion-aware spatial attention module.
Our main contributions are summarized as follows.

\begin{itemize}
\setlength{\itemsep}{0pt}
\setlength{\parsep}{0pt}
\setlength{\parskip}{0pt}
  \item To the best of our knowledge, we are the \textit{first} to embed motion information into an end-to-end VOS pipeline with a single branch and without a precomputed optical flow.
  \item We introduce the MU-Layer and MSAM to learn the motion features with uncertainty, which provides critical information for VOS.
  \item Experimental evaluation on benchmark datasets with wide protocols verifies the superiority of our proposed method, where our method is competitive with the existing \textit{SOTA} methods, especially the best performance under low-data protocols, \ie (1), (3-1).

\end{itemize}

\section{Related Work}
In this section, we introduce previous works from two categories of VOS as feature matching based methods and motion based methods. 

\noindent{\textbf{Feature matching methods.}}
STM\cite{oh_STM_ICCV_2019} achieves great success, which performs dense feature matching across the entire spatio-temporal volume of the video through a dynamic memory bank, \ie saves the spatio-temporal information of previous frames. 
To alleviate the problem of missing samples and out-of-memory crashes when processing long videos, AFB-URR\cite{liang_AFB_URR_NIPS_2020} introduces the adaptive feature bank to organize the object features by weighted averaging and discards obsolete features by least frequently used index principle. 
STM-Cycle\cite{Li_STM-cycle_NeurIPS_2020} incorporates cycle consistency to mitigate the error propagation problem. 
\cite{liyu_fast_vos_ECCV_2020} proposes a global context module using attentions to reduce temporal information and guide the segmentation.
RANet\cite{Ziqin_RANet_CVPR_2019} is a hybrid strategy, integrating the insights of matching based and propagation based methods to learn pixel-level similarity. 
KMN\cite{seong_kernelizedvos_eccv_2020} and GraphMem\cite{lu_episodicvos_eccv_2020} focus more on memory bank optimization to achieve more accurate key-value matching, both are accompanied by data augmentation, such as Hide-and-Seek and Label Shuffling, and their performance will drop a lot if the augmentation is removed.
We agree that these schemes will also give us an improvement, but this paper is dedicated to using simple motions to emphasize the essentials.
Besides, although these methods have used temporal information to improve accuracy, they ignore the most essential motion information.

\noindent{\textbf{Motion-based methods.}}
Such methods can be roughly classified as mask propagation methods \cite{Perazzi_masktrack_CVPR_2017, oh_seoung_fastvos_RGMP_CVPR_2018,li_vsreid_cvprw_davis_2017, xie_RMNet_CVPR_2021} and two-stream methods \cite{cheng_segflow_cvpr_2017, khoreva_lucid_dream_DAVIS_2017, Dave_seg_anything_moves_ICCV_2019,jang_online_vos_ctn_cvpr_2017} . 
The mask propagation methods start from an annotated frame and propagates masks through the entire video sequence, and the propagation process is often guided by optical flow. 
Two-stream architecture usually fuses feature between appearance (RGB branch) and motion (optical flow branch). 

However, such methods highly dependent on high-quality pre-inferenced optical flow \cite{ilg_flownet2_cvpr_2017,hui_liteflownet_cvpr_2018,revaud_epicflow_cvpr_2015, hu_efficient_flow_cvpr_2016, kroeger_fast_flow_eccv_2016, teed_raft_eccv_2020}, and the performance will be limited by the generalization ability of optical flow network due to the fact that there is no ground truth of optical flow in the common VOS task.
RMNet \cite{xie_RMNet_CVPR_2021} uses the mask of the previous frame and pre-computed optical flow to generate the attention mask for current frame, but different optical flow networks \cite{ilg_flownet2_cvpr_2017, teed_raft_eccv_2020} are selected for different datasets. It also shows that this kind of methods cannot escape the disadvantages of generalization.
As for mask propagation methods, they rely on temporal continuity from optical flow and spatio-temporal context from the previous frames, this leads these methods are difficult to deal with occlusions, rapid motion, and complex deformation of objects, also meets performance drift over time once the propagation becomes unreliable.
\section{Motion Uncertainty-aware VOS}

Given a video sequence with ${T}$ frames, $\mathcal{I}_{t} \in \mathbb{R}^{H \times W \times 3}$, $t \in [1,T]$ is the $t$-th frame RGB image, $\mathcal{M}_{t} \in \mathbb{R}^{H \times W \times 1}$ is the ground truth object segmentation mask and $\widehat{\mathcal{{M}}}_t$ denotes the predicted object mask. 
Semi-supervised VOS aims to predict the object masks of all following video frames with the first RGB image $\mathcal{I}_{1}$ and its object annotation mask $\mathcal{M}_{1}$ as prior, which can be formulated as 
${\widehat{\mathcal{M}}_t} = 
        \mathcal{H}_{\theta} ([\mathcal{I}_{1}, \mathcal{M}_{1}...\mathcal{I}_{t-1}, \widehat{\mathcal{M}}_{t-1}], \mathcal{I}_{t})$,
where $\mathcal{H}_{\theta}$ is an object segmentation network with learnable weights $\theta$. $[\mathcal{I}_{1}, \mathcal{M}_{1}...\mathcal{I}_{t-1}, \widehat{\mathcal{M}}_{t-1}]$ denotes the historical frames before the current frame $t$ that can be used directly or implicitly to infer the object mask.

\begin{figure*}[!ht] 
    \centering
    \includegraphics[width=\textwidth]{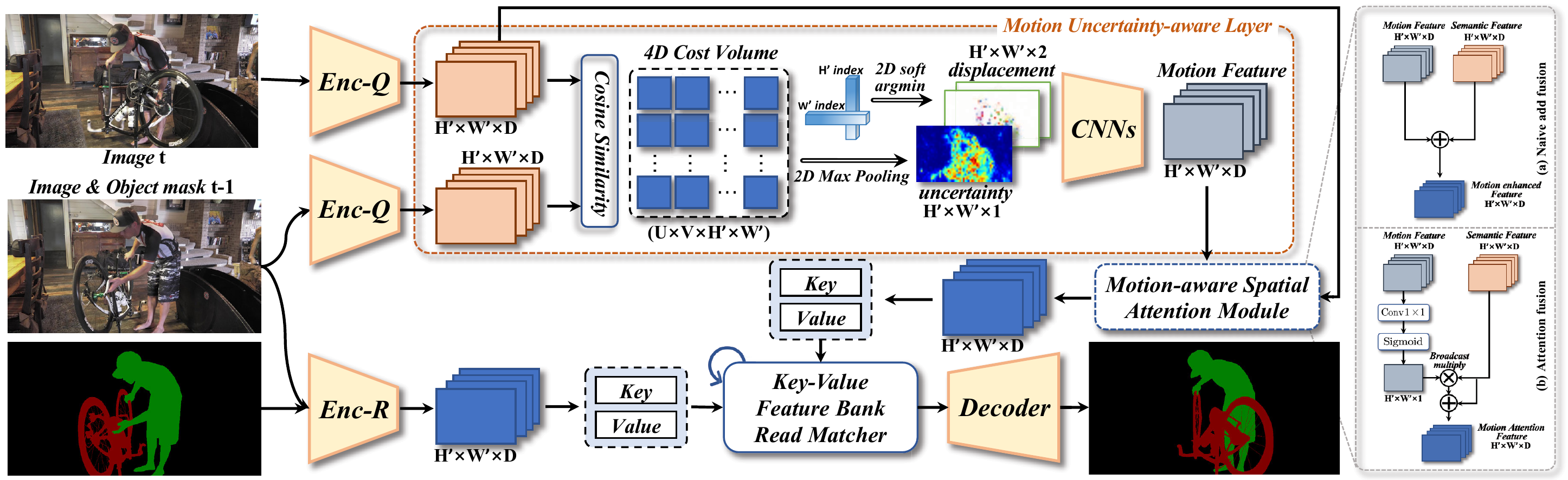}
    \vspace{-\baselineskip}
    \caption{\textbf{MUNet architecture.}
    First, we use \textbf{\textit{Enc-Q}} and \textbf{\textit{Enc-R}} to extract the semantic features $\mathcal{F}^{s}_{t}, \mathcal{F}^{s}_{t-1}$ of $\mathcal{I}_t, \mathcal{I}_{t-1}$ from image appearance. 
    Then, we use \textbf{MU-Layer} to calculate the motion feature $\mathcal{F}^{m}_{t}$, and feed it into \textbf{MSAM} to guide and enhance the semantic feature $\mathcal{F}^{s}_{t}$. The matching results of Key-value Matcher is fed into the \textbf{Decoder} to generate the object mask $\widehat{\mathcal{M}}_t$, and the memory feature-bank is updated dynamically over time.}
    \label{fig:OurFramework}
\end{figure*}

\subsection{Network Overview}
Our MUNet is a pixel-level dense matching method. The network architecture as illustrated in \figref{fig:OurFramework} consists of four seamless parts: 1) two encoders as feature extractors, 2) a Motion Uncertainty-aware Layer (MU-Layer), 3) a Motion-aware Spatial Attention Module (MSAM), and 4) a decoder with a feature memory bank.

First, the reference image $\mathcal{I}_{t-1}$ and mask $\mathcal{M}_{t-1}$ are fed into the reference encoder \textit{Enc-R}, the query image $\mathcal{I}_{t}$ and adjacent frame $\mathcal{I}_{t-1}$ are respectively input to query encoder \textit{Enc-Q} with shared weights.
Second, the semantic features $\mathcal{F}^{s}_{t-1}, \mathcal{F}^{s}_{t}$ from \textit{Enc-Q} are input to the MU-Layer to obtain the motion feature $\mathcal{F}^{m}_{t}$, more details are described in \secref{sec::uncertainty_aware_motion_rep}.
Then, we use MSAM to take advantage of the motion feature $\mathcal{F}^{m}_{t}$ from adjacent frames to enhance the semantic feature embed by \textit{Enc-Q} in \secref{sec::motion_attention}. 

In \cite{oh_STM_ICCV_2019, Li_STM-cycle_NeurIPS_2020, liang_AFB_URR_NIPS_2020}, the semantic features output by \textit{Enc-R} and \textit{Enc-Q} are directly used for key-value embedding, where the keys are used for addressing and matching, while the values are used to preserve feature information. 
This will cause the query process to rely heavily on the similarity matching of appearance semantic features, which may lead to appearance confusion and wrong predictions caused by similar instances. 
Different from them, for a query image, we use the motion enhanced feature from MASM for key-value embedding, and obtain the key-value pairs $\{k_t^Q, v_t^Q\}$ to match the most similar features in the memory bank.
The key-value pairs $[\{k_1^R, v_1^R\},...,\{k_{t-1}^R, v_{t-1}^R\}]$ of historical images are stored in memory bank, updated dynamically over time.
Finally, the matching results from memory bank are fed into decoder, and then output the segmentation mask of each object.
We use ResNet50\cite{kaiming_resnet50_cvpr_2016} as the backbone of two encoders. 
For the query encoder \textit{\textbf{Enc-Q}}, we take the output feature map of Layer-4 (\textit{res4}) as a semantic feature $\mathcal{F}^{s}_{t} \in \mathbb{R}^{(H/16) \times (W/16) \times 1024}$, where $H$ and $W$ are the height and width of raw image shape correspondingly. 
The reference encoder \textit{\textbf{Enc-R}} is slightly different from the vanilla ResNet50, which takes the RGB image (3-channel) and its segmentation label mask (1-channel) as inputs to extract object-level semantic feature.

\subsection{Motion Uncertainty-aware Layer (MU-Layer)}
\label{sec::uncertainty_aware_motion_rep}

Here we introduce MU-Layer, which establishes spatio-temporal relationships based on the motion feature calculated from high-level semantic features of adjacent frames. 
We construct a cost volume to model pixel-wise feature similarity that indicates the rough motion.

\noindent\textbf{Cost Volume.}
For semantic features $\mathcal{F}^{s}_{t-1}, \mathcal{F}^{s}_{t}\in \mathbb{R}^{H^{\prime}\times W^{\prime}\times D}$ of reference and query images from \textit{Enc-Q}, where $H^{\prime}, W^{\prime}$ indicates ${H}/{16}, {W}/{16}$ of the input image shape, $D$ is the feature dimension. 
In order to achieve lightweight computing, we first reduce the feature dimension to ${D}/{4}$ by $1 \times 1$ convolution.
Then we define the correlation as ${\rm{\mathbf{C}}} \in \mathbb{R}^{U\times V\times H^{\prime}\times W^{\prime}}$, which can be efficiently computed as cosine similarity between each pixel in the $H^{\prime}\times W^{\prime}$ reference image with a set of candidate targets in a $U \times V$ search window. 

\begin{equation} 
\label{equa:corr}
\begin{aligned}
    {\rm{\mathbf{C}}}({\rm \mathbf{u}}, {\rm \mathbf{x}})
    = \frac{\mathcal{F}^{s}_{t}({\rm \mathbf{x}}) \bigcdot \mathcal{F}^{s}_{t-1}({\rm \mathbf{x}}+{\rm \mathbf{u}})}
    {
     \big|\big| \mathcal{F}^{s}_{t}({\rm \mathbf{x}}) \big|\big| \bigcdot
     \big|\big| \mathcal{F}^{s}_{t - 1}({\rm \mathbf{x}}+{\rm \mathbf{u}}) \big|\big|
    },
\end{aligned}
\end{equation}
where ${\rm \mathbf{x}}\!=\!(x,y)$ is the source pixel coordinate, ${\rm \mathbf{u}}\!=\!(u,v)$ is the pixel displacement in search window. We set window size as 25$\times$25.

\noindent\textbf{Displacement Calculation.}
In order to \textit{implicitly represent the motion without optical flow ground truth as supervision},
we dig into the information represented by cost volume and calculate the displacement with the highest matching cost.
We use a soft-argmin operator by $\widehat{{\rm \mathbf{u}}}=\sum_{{\rm \mathbf{u}} \in \mathcal{U}}[{\rm \mathbf{u}} \times \sigma({-\rm \mathbf{C{(u, x)}}})]$ to solve these problems. $\sigma$ denotes the softmax function. We use 2D softmax along the displacement hypothesis space of ${\rm \mathbf{u}}$ to calculate the matching probability. Therefore, $\widehat{\mathbf{u}}$ is a two-channel displacement vector.

\noindent\textbf{Uncertainty Estimation.}
The last two dimensions of the cost volume represent the correspondence between $\mathcal{F}^{s}_{t}({\rm \mathbf{x}})$ and $\mathcal{F}^{s}_{t-1}({\rm \mathbf{x}}+{\rm \mathbf{u}})$ in all spatial positions. 
Previously, we build pixel-wise displacement between features of adjacent frames by soft-argmin, but it may fail in low texture, large motion, and motion blur areas.
To remedy this issue, we propose an uncertainty branch to measure the matching confidence.
We use the matching cost value of the correlation matrix in each spatial position as a score to measure the matching uncertainty of each pair of spatial correspondence.
Specifically, we use max-pooling to calculate the max value of coordinate $(x, y)$ as matching uncertainty, which means the highest response in each spatial position of 4D cost volume.
We define uncertainty map ${\rm U_{map}} \in \mathbb{R}^{H^{\prime} \times W^{\prime} \time 1}$ as,

\begin{equation}
\label{equa:uncertainty_map}
\begin{aligned}
     {\rm U_{map}}(x,y) = \mathop{\max}\limits_{\{u,v\}} {\mathbf{C}}((u,v), (x,y)).
\end{aligned}
\end{equation}

\begin{figure}[!h]
    \centering
    \includegraphics[width=0.72\linewidth]{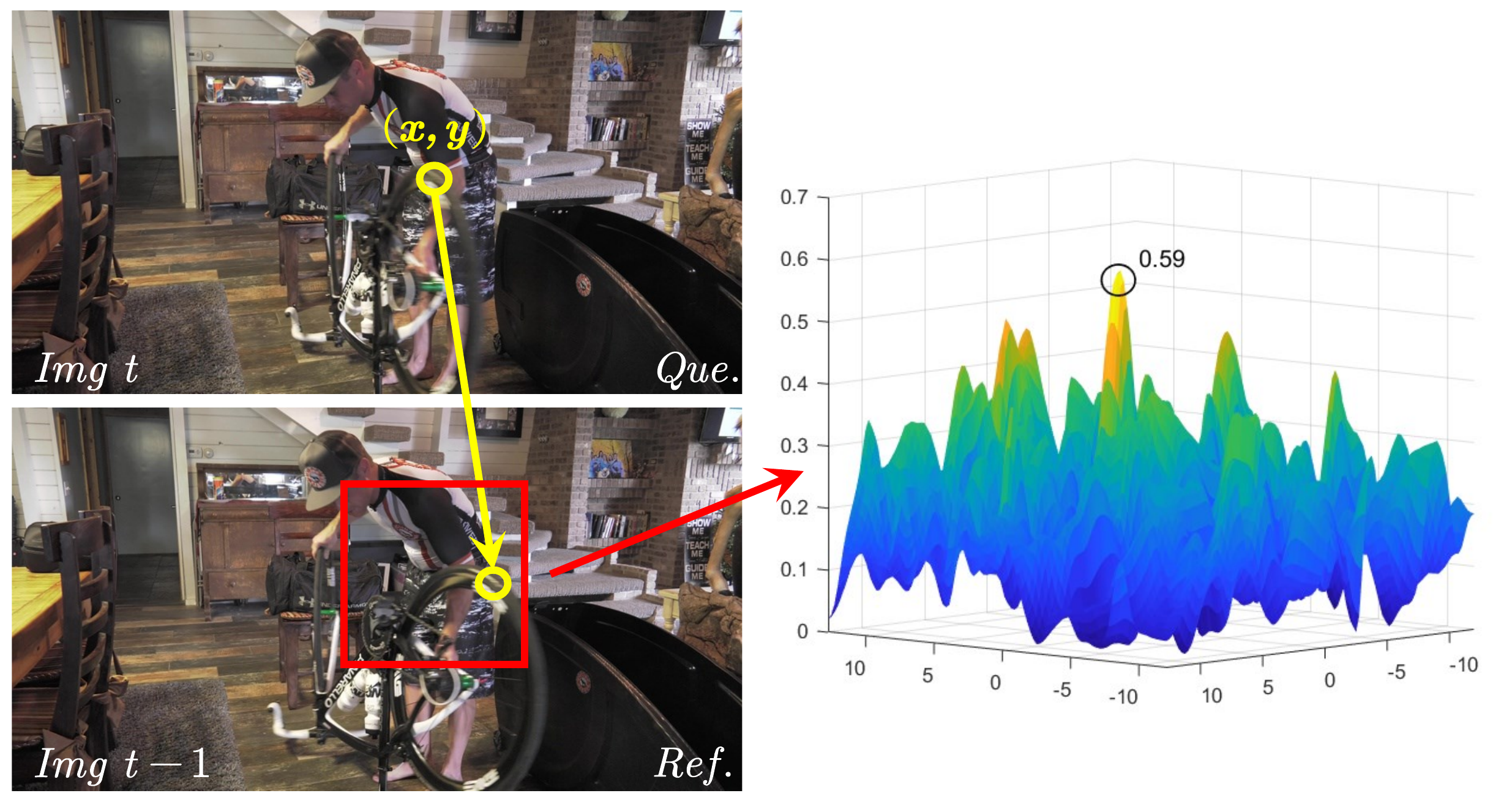}
    \caption{\textbf{Uncertainty map calculation.}}
    \label{fig:danfeng}
\end{figure}

As shown in \figref{fig:danfeng}, we randomly select a pixel $(x, y)$ in query image $\mathcal{I}_{t}$, the red box is the corresponding search window in reference image $\mathcal{I}_{t-1}$, and yellow line represents the correspondence between two adjacent frames. Right subfigure indicates the local matching cost of the search window, which will be spliced into the total uncertainty map ${\rm U_{map}^t}$ of image $\mathcal{I}_t$ as \equref{equa:uncertainty_map}.

\noindent\textbf{Motion Feature Representation.}
After getting motion information from cost volume by soft-argmin and max-pooling, we also need a high-level motion feature to achieve feature fusion with high-level semantic feature on an equivalent level.
Different from many two-stream based VOS methods\cite{li_MotionAttention_ICCV_2019, zhou_matnet_aaai_2020_tip} that directly use a deep encoder parallel with RGB encoder to encode motion feature, we design a \textit{lightweight CNN model} to project the original motion information ($H^{\prime} \times W^{\prime} \times 3$) into a high-dimensional feature space ($H^{\prime} \times W^{\prime} \times D$). How to use the motion feature to enhance the semantic feature will be described in \secref{sec::motion_attention}. \tabref{table:motion_net} shows the structure of our \textit{lightweight} motion feature representation network in our proposed MU-Layer, which maps the original motion information ($H^{\prime} \times W^{\prime} \times 3$) into a high-dimensional feature space ($H^{\prime} \times W^{\prime} \times D$).

\begin{table}[!ht]
    \renewcommand\arraystretch{0.9}
    \setlength\tabcolsep{4pt}
    \centering
    \resizebox{0.5\linewidth}{!}{
    \begin{tabular}{c|rrrr}
    \toprule
    {Layer} & $C_{in}$, &  $C_{out}$, & $kernel$, & $stride$ \\ \midrule
    1     & 3, & 3, & 7, & 1 \\
    2     & 3, & 16,& 1,& 1 \\
    3     & 16,& 16,& 3,& 1 \\
    4     & 16,& 32,& 1,& 1 \\
    5     & 32,& 32,& 3,& 1 \\
    6     & 32,& 32,& 1,& 1 \\
    7     & 32,& 64,& 3, &1 \\
    8     & 64,& 1024,& 1,& 1 \\
    \toprule
    \end{tabular}
    }
    \caption{\textbf{Structure of the motion feature representation network}. We uses 2D convolution for lightweight computing, instead of using 3D convolution.}
    \label{table:motion_net}
\end{table}
\vspace{-\baselineskip}

\subsection{Motion-aware Spatial Attention Module (MSAM)}
\label{sec::motion_attention}
To enhance the semantic feature from \textit{Enc-Q} by motion feature provided by the former MU-Layer, a direct and simple way is fusing the motion feature with semantic feature by element-wise add operation, as shown in the right of \figref{fig:OurFramework}(a) and can be formulated as $\mathcal{F}^{out} = \mathcal{F}^s + \mathcal{F}^m$,
where $\mathcal{F}^m , \mathcal{F}^s$ denotes the motion and semantic feature respectively, $\mathcal{F}^{out}$ is the fused feature, which is subsequently used to embed the key-value pair of query image.

Considering that the motion information is different in each spatial position, to use it more efficiently, we design \textit{a lightweight attention module} to fuse the motion feature and semantic feature. 
In this way, the semantic feature from \textit{Enc-Q} can be enhanced by the motion feature in some important regions of motion.
Thus, we exploit motion feature as spatial attention weights, as shown in \figref{fig:OurFramework}(b), and this module can be formulated as,

\begin{equation} 
\label{equa:attention}
\begin{aligned}
    \mathcal{F}^{out} = \mathcal{F}^s \otimes {\rm sigmoid}({\rm Conv_{1 \times 1}}(\mathcal{F}^m)) + \mathcal{F}^s,
\end{aligned}
\end{equation}
\noindent where $\otimes$ indicates the broadcast multiply, the shape of $\mathcal{F}^m$ is $H^{\prime} \times W^{\prime} \times D$ and the output channel of ${\rm Conv_{1 \times 1}}$ is $1$.
So after ${\rm Conv_{1 \times 1}}$ and ${\rm sigmoid}$, we can get one-channel attention map with shape $H^{\prime} \times W^{\prime} \times 1$. 

\subsection{Loss function}
Cross-Entropy (CE) is most widely used in segmentation tasks.
However, the CE loss calculates the error of each pixel independently and treats all pixels equally, which ignores the global structure of the image. 
Here we adopt the Bootstrap Cross-Entropy loss (BsCE, $\mathcal{L}_{bsce}$) \cite{wu_bridging_bootstrap_celoss_2016} to force networks to focus on the hard and valuable pixels during training, and we select top $40\%$ hardest pixels to carry out back propagation.
Besides, we use the mask-IoU loss ($\mathcal{L}_{miou}$) to optimize the global structure instead of focusing on a single independent pixel, which is not affected by the unbalanced distribution. 
Thus, we use the combination of $\mathcal{L}_{bsce}$ and $\mathcal{L}_{miou}$ as supervision, which is formulated as,

\begin{gather}
\label{equa:loss_function}
\begin{aligned}
 \mathcal{L}{(\widehat{\mathcal{M}}_t, \mathcal{M}_t)} &= \mathcal{L}_{bsce} \!+\! \lambda \mathcal{L}_{miou} \\
 & = \frac{1}{\phi(\Omega)} \!\sum_{p \in \phi(\Omega)}\!((1\!-\!\mathcal{M}_{t}^{p})\log(1\!-\!\widehat{\mathcal{M}}_{t}^{p})\!+\!\mathcal{M}_{t}^{p}\log(\widehat{\mathcal{M}}_{t}^{p}))
\\& \quad + \lambda \left[ 1-\frac{\sum_{p \in \Omega} \min(\mathcal{\mathcal{M}}_{t}^{p}, \widehat{\mathcal{M}}_{t}^{p})}{\sum_{p \in \Omega} \max(\mathcal{M}_{t}^{p}, \widehat{\mathcal{M}}_{t}^{p})} \right],
\end{aligned}
\raisetag{25pt}
\end{gather}
\noindent where $\lambda$ is a trade-off parameter and we set $\lambda=1$ in all experiments. $\Omega$ is the set of all pixel in the mask, and $\phi(\Omega)$ is the hardest region used for bootstrap CE loss. $\widehat{\mathcal{M}}_{t}$ denotes the predicted object mask and $\mathcal{M}_{t}$ is the ground truth. 
We demonstrate the advantages of this combination of loss function in \tabref{table:ablation_study}.


\subsection{Implementation Details}
We implement our model by PyTorch \cite{paszke_pytorch_NIPS_2019} with a single NVIDIA RTX 2080Ti GPU.
We use ResNet50 \cite{kaiming_resnet50_cvpr_2016} as the feature extractor (\textit{Enc-Q, Enc-R}), pretrained on ImageNet. 
During training, we select 6 continuous frames per sequence as a batch (one as reference frame and the other five as query frames). 
During inference, we feed the video sequence frame by frame and do not use online fine-tuning.
We simply apply common data augmentation on current frames including flip, color jitter and affine transformation.
The input frames are randomly resized and cropped into 400 $\times$ 400, and we use the raw image size during inference. 
We minimize $\mathcal{L}_{bsce}$ and $\mathcal{L}_{miou}$ by the AdamW \cite{loshchilov_AdamW_ICLR_2018} optimizer with default parameters $\beta_1=0.9, \beta_2=0.999$. The initial learning rate is $2 \times 10^{-5}$ and the weight decay is $0.5$.
\section{Experiments}

\subsection{Datasets, Evaluation metrics, and Protocols}
We train and evaluate our method on DAVIS 2017 \cite{Pont_DAVIS_arxiv_2017} and YouTube-VOS 2018 \cite{Xu_YouTube_ECCV_2018} datasets. 
Considering that many methods use extra static image datasets for pre-training \cite{Cheng_PAMI_2015,everingham_pascal_voc_2010,shi_hierarchical_pami_2015,lin_coco_microsoft_2014,li_secrets_2014} or conduct fine-tuning, for a fair comparison, we categorize and compare solutions based on whether they use extra static datasets and online fine-tuning.

\noindent\textbf{DAVIS 2017.}
The DAVIS17 dataset contains 120 video sequences in total, where 60 sequences are split for training, 30 for validation, and 30 for testing. Each video contains one or several annotated objects to track. Each video sequence has 25 to 104 annotated continuous frames.

\noindent\textbf{Youtube-VOS 2018.}
The Youtube-VOS18 dataset contains 4453 videos with one or more target objects, including 3471 videos for training (65 categories), 474 sequences for validation (additional 26 unseen categories).
Each video sequence has 20 to 180 discontinuous frames, where every 5 interval frames are provided and annotated. 

\noindent\textbf{Evaluation metrics.}
Following the standard DAVIS protocol, we measure region accuracy $\mathcal{J}$ by calculating average intersection-over-union (IoU), and boundary accuracy $\mathcal{F}$ via bipartite matching between boundary pixels.
In addition, we compare inference speed by frames per second (FPS) according to averaging FPS of each sequence on the validation set.

\begin{figure}[!ht]
    \centering
    \includegraphics[width=0.62\linewidth]{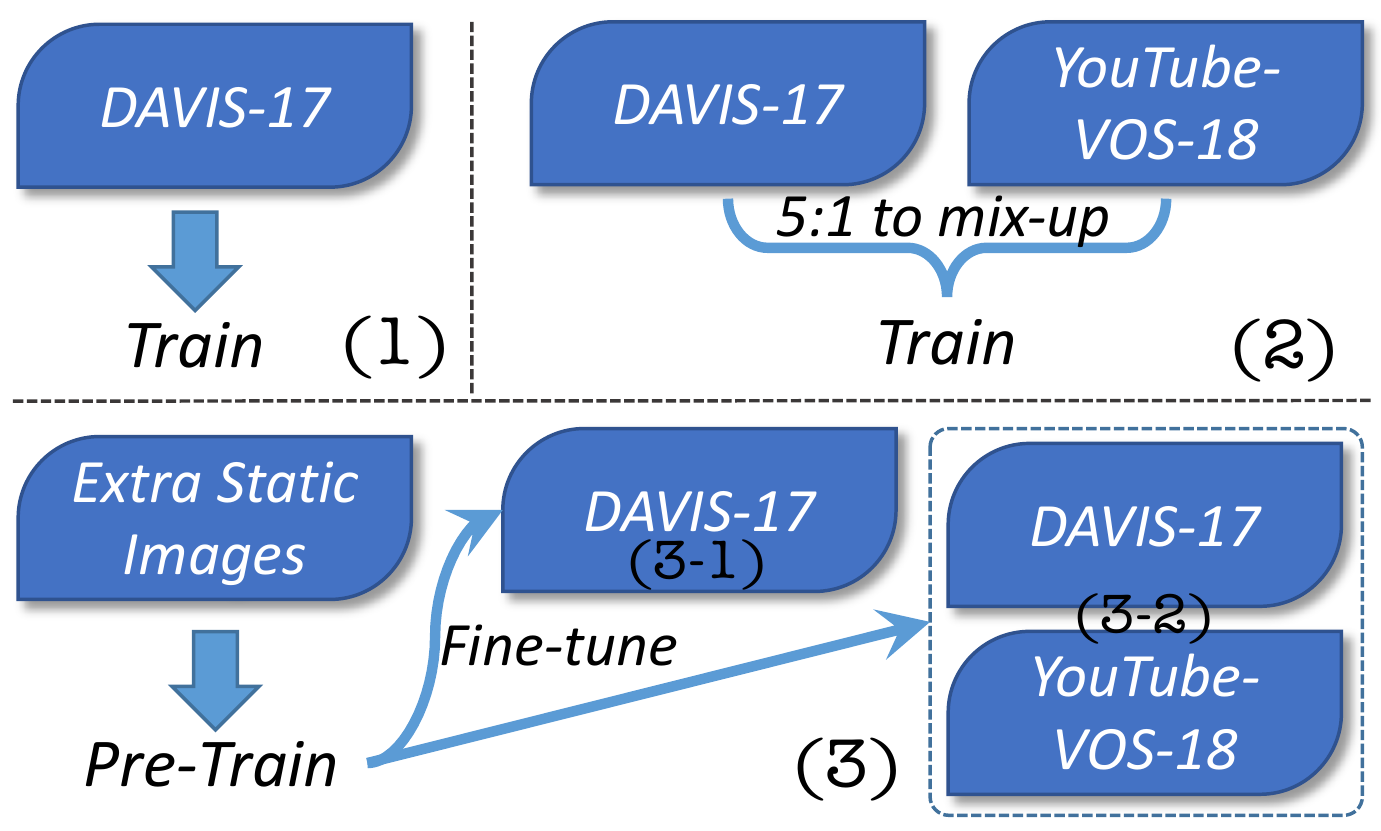}
    \vspace{-10pt}
    \caption{Dataset usage protocols.}
    \label{fig:train_strategy_no_warp}
\end{figure}

\noindent\textbf{Evaluation protocols.} There are various training protocols for the existing VOS methods. For a fair comparison with more methods, we train our model and report the results under different data use protocols, including \texttt{(1)} only using DAVIS17 for training, \texttt{(2)} jointly training on DAVIS17 and YouTube-VOS18, \texttt{(3)} using static images \cite{Cheng_PAMI_2015,everingham_pascal_voc_2010,shi_hierarchical_pami_2015,lin_coco_microsoft_2014,li_secrets_2014} to pretrain and then fine-tuning on DAVIS17 \texttt{(3-1)} or both jointly \texttt{(3-2)}, more intuitively in \figref{fig:train_strategy_no_warp}.
The different training settings may cause unfair comparisons, such as using different backbones (\eg RestNet50, ResNet101), using well-designed data augmentation methods (\eg. Hide-and-Seek, Label Shuffling, Balanced Random-Crop), using different weight initialization (\eg Mask-RCNN, DeepLab) and whether performing online fine-tune on the test video or not, \etc.
Here, we will report the detailed comparisons as fair as possible. 

It is worth mentioning that, this low-data experimental setup \textit{greatly reduces the training time and no longer requires a large amount of data to pretrain}. 
Although this is a smart way to initialize the network and improve the $\mathcal{J} \& \mathcal{F}$ metric under \texttt{protocol(3)}, researchers gradually realize the shortcomings of this way, such as long training time. 
In our setting, one epoch of pretraining under \texttt{protocol(3)} needs about 17 hours, and the whole pretraining process may take 6$\sim$7 days in a single 2080Ti GPU.
The time of one epoch under \texttt{protocol(2)} is about 25 minutes and \texttt{protocol(1)} is about 12 minutes. 
The training \texttt{protocol(1)} and \texttt{(2)} are lighter, \ie less time.
We believe abandoning the pretraining phase by emphasizing own data motion information is an improvement in the VOS field.

\subsection{Quantitative Comparison} \label{sec::sota_comparison}

\textbf{DAVIS17.} Results on the DAVIS17 validation set are reported in \tabref{table:result_davis_17}. 
Our method achieves \textit{\textbf{1st}} under training \texttt{protocol}\texttt{(1)} and \texttt{protocol}\texttt{(3-1)}, and ranks \textit{\textbf{2nd}} under \texttt{protocol(2)}.
Under \texttt{protocol(1)}, only using the DAVIS17 dataset for training, our vanilla version achieves $\mathbf{75.0\%}$ on $\mathcal{J}\&\mathcal{F}$, which outperforms all other methods. 
And a better initialization by Mask-RCNN ResNet50 can boost the performance of our method to $\mathbf{76.5\%}$ on $\mathcal{J}\&\mathcal{F}$, which significantly outperforms STM\footnotesize($43.0\%$)\normalsize \ by $36.5\%$ and CFBI\footnotesize($74.9\%$)\normalsize \ by $1.4\%$.
In addition, when jointly training under \texttt{protocol(2)}, ours achieves $\mathbf{81.1\%}\ \mathcal{J}\&\mathcal{F}$, and the performance can increase to $\mathbf{81.9\%}$ if using static images to pretrain. 
Ours surpasses most competitors in \tabref{table:result_davis_17}, it is competitive and able to prove our superiority.

\begin{table}[!ht]
    \renewcommand\arraystretch{0.92}
    \setlength\tabcolsep{4pt}
    \centering
    \resizebox{0.75\linewidth}{!}{
    \begin{tabular}{ll|cc|ccc|c}
    \toprule
    \multicolumn{2}{c|}{\textbf{Methods}} & \textbf{S} & \textbf{Y} & $ \bm{\mathcal{J}}$ & $\bm{\mathcal{F}}$  & $\bm{\mathcal{J}$\&$\mathcal{F}}$ & \textbf{FPS} \\ 
    \toprule
    \multirow{8}{*}{\rotatebox{270}{Protocol (1)}} 
    & STM\footnotemark[1] \cite{oh_STM_ICCV_2019}   &  &  & - & - & 43.0 & 6.25\\
    & OSMN \cite{yang_osmn_cvpr_2018}   &  &  & 52.5 & 57.1 & 54.8 & -\\
    & OnAVOS \cite{voigtlaender_OnAVOS_BMCV_2017}   &  &  & 64.5 & 71.2 & 67.9 & 0.08\\
    & FRTM\cite{Robinson_frtm-vos_CVPR_2020}    &  &  & -    & -    & 68.8 & 21.9 \\
    & FEELVOS\cite{voigtlaender_feelvos_CVPR_2019}  &  &  & 65.9 & 72.3 & 69.1 & 2.22 \\
    & LWLVOS$^{\ddagger}$\cite{bhat_learn_what_vos_ECCV_2020} &  &  & 72.2  & 76.3   & 74.3  & - \\
    & CFBI$^{\dagger}$ \cite{yang_CFBI_ECCV_2020}   &  &  & 72.1 & 77.7 & 74.9  & 5.55 \\
    & \textbf{Ours} & &  & \underline{72.4}   & \underline{77.8}   & \underline{75.0} & 6.42 \\ 
    & \textbf{Ours-v2}$^{\ddagger}$ & &  & \textbf{74.5} & \textbf{78.5} & \textbf{76.5} & 6.42 \\ 
    
    \midrule \midrule
    \multirow{7}{*}{\rotatebox{270}{Protocol (2)}} 
    & A-GAME\cite{Johnander_AGame_CVPR_2019}        &  & \checkmark & 67.2   & 72.7   & 70.0 & 14.3 \\
    & FEELVOS\cite{voigtlaender_feelvos_CVPR_2019}  &  & \checkmark & 69.1   & 74.0   & 71.5 & 2.22 \\
    & STM-cycle\cite{Li_STM-cycle_NeurIPS_2020}     &  & \checkmark & 69.3   & 75.3   & 72.3 & 9.3 \\
    & AFB-URR$^{*}$\cite{liang_AFB_URR_NIPS_2020}     &  & \checkmark  & 72.0 & 75.7 & 73.9 & - \\
    & PMVOS\cite{cho_pmvos_arxiv_2020}              &  & \checkmark & 71.2   & 76.7   & 74.0 & -\\
    & FRTM\cite{Robinson_frtm-vos_CVPR_2020}    &  & \checkmark & -      & -      & 76.7 & 21.9 \\
    & CFBI$^{\dagger}$ \cite{yang_CFBI_ECCV_2020}   &  & \checkmark  & \textbf{79.1}   & \textbf{84.6}   & \textbf{81.9}  & 5.55 \\
    & \textbf{Ours} & & \checkmark & \underline{78.3} & \underline{83.9} & \underline{81.1} & 6.42\\ 

    \midrule \midrule
    \multirow{11}{*}{\rotatebox{270}{Protocol (3-1)}} 
    & RANet\cite{Ziqin_RANet_CVPR_2019} & \checkmark &  & 63.2  & 68.2   & 65.7 & 30.3\\
    & RGMP\cite{oh_seoung_fastvos_RGMP_CVPR_2018} & \checkmark &  & 64.8  & 68.6 & 66.7 & 7.70\\
    & OSVOS$^{\rm{S}}$ \cite{maninis_OSVOS-S_CVPR_2018} & \checkmark &  & 64.7  & 71.3 & 68.0 & 0.22\\
 
    & CINM\cite{bao_cinm_cvpr_2018}          & \checkmark &  & 67.2  & 74.2   & 70.7 & -\\
    & GC\cite{li_gc_fast_eccv_2020}             & \checkmark &  & 69.3  & 73.5   & 71.4 & - \\

    & STM\cite{oh_STM_ICCV_2019}             & \checkmark &  & 69.2  & 74.0   & 71.6 & 6.25 \\
    & AFB-URR\cite{liang_AFB_URR_NIPS_2020}  & \checkmark &  & 73.0  & 76.1   & 74.6 & 6.18 \\
    & {RMNet\cite{xie_RMNet_CVPR_2021}}  & {\checkmark} &  & {72.8}  & {77.2} & {75.0} & {-}\\
    & LCM\cite{hu_lcm_cvpr_2021}             & \checkmark &  & 73.1  & 77.2   & 75.2 & - \\
    & PReMVOS\cite{luiten_premvos_accv_2018}  & \checkmark &  & 73.9  & \textbf{81.7} & \underline{77.8} & 0.03\\ 
    & KMN\cite{seong_kernelizedvos_eccv_2020} & \checkmark &  & \underline{74.2}  & 77.8 & 76.0 & 8.33 \\ 
    & \textbf{Ours} & \checkmark &  & \textbf{74.7}   & \underline{81.4}& \textbf{78.1} & 6.42 \\ 
    
    \midrule \midrule
    
    \multirow{5}{*}{\rotatebox{270}{(3-2)}} 
    & STM\cite{oh_STM_ICCV_2019} & \checkmark & \checkmark & 79.2  & 84.3 & 81.8 & 6.25 \\
    & GMVOS\cite{lu_episodicvos_eccv_2020} & \checkmark & \checkmark & {80.2}  & {85.2} & \underline{82.8} & 5.00 \\
    & KMN\cite{seong_kernelizedvos_eccv_2020} & \checkmark &  \checkmark & {80.0} & {85.6} & \underline{82.8} & 8.33 \\ 
    & {RMNet\cite{xie_RMNet_CVPR_2021}} & {\checkmark} & {\checkmark} & {\textbf{81.0}} & {\underline{86.0}} & {\textbf{83.5}} & {-} \\ 

    & LCM\cite{hu_lcm_cvpr_2021} & \checkmark & \checkmark & \underline{80.5}  & \textbf{86.5}  & \textbf{83.5} & - \\
    & \textbf{Ours} & \checkmark & \checkmark & {79.2} & {84.6} & {81.9} & 6.42\\ 
    \toprule
    \end{tabular}
    }
    \caption[Comparison for DAVIS]{{\textbf{Comparison on the DAVIS17 validation set.}}{\footnotemark[1]}}  \label{table:result_davis_17}
\end{table}

\footnotetext[1]{In \tabref{table:result_davis_17},\ref{table:result_youtube_18}, ``\textbf{S}'': static images for training, ``\textbf{Y}'': Youtube-VOS18, and ``\textbf{O}'': fine-tune on test strategy. $^{\dagger}$ indicates using DeepLabv3-ResNet101 to initialize the backbone network, and $^{\ddagger}$ means using Mask-RCNN-ResNet50. 
The best result is \textbf{bold-faced}, and the suboptimal result is \underline{underlined}. 
}

\noindent\textbf{YouTube-VOS18}.
We report the results on the YouTube-VOS18 validation set in \tabref{table:result_youtube_18}. 
The subscript $\mathcal{U}$ and $\mathcal{S}$ denote the unseen and seen categories, $\mathcal{G}$ is the average of all four measures. Our method with different training settings can achieve competitive results.
Compared with the baseline STM\cite{oh_STM_ICCV_2019}, the motion information brings $0.9\%$  \footnotesize($79.4\%$$\rightarrow$$80.3\%$)\normalsize \ performance gain. Especially without pretraining on static images, our mehtod brings $9.2\%$  \footnotesize($68.2\%$$\rightarrow$$77.4\%$)\normalsize \ gain. In addition, we found the improvement on YouTube is not as significant as DAVIS. The reason is that DAVIS consists of continuous frames, while every 5 interval frames (discontinuous) are used in YouTube. MULayer cannot effectively capture the long range motion information without the optical flow ground truth supervision.

\begin{table}[!ht]
    \renewcommand\arraystretch{0.92}
    \setlength\tabcolsep{4pt}
    \centering
    \resizebox{0.75\linewidth}{!}{%
        \begin{tabular}{l|cc|cccc|c}
        \toprule
        \textbf{Methods} & \textbf{S} & \textbf{O} & $\bm{\mathcal{J_U}}$ & $\bm{\mathcal{F_U}}$ & $\bm{\mathcal{J_S}}$ & $\bm{\mathcal{F_S}}$ & $\bm{\mathcal{G}}$  \\ 
        \toprule
        S2S\cite{Xu_YouTube_ECCV_2018} & & \checkmark & 55.5 & 61.2 & 71.0 & 70.0 & 64.4 \\
        PReMVOS\cite{luiten_premvos_accv_2018}&  & \checkmark & 56.6 & 63.7 & 71.4 & 75.9 & 66.9 \\
        STM-cycle\cite{Li_STM-cycle_NeurIPS_2020}&  &  \checkmark & 62.8 & 71.9  & 72.2 & 76.3 & 70.8 \\
        MSK\cite{Perazzi_masktrack_CVPR_2017} & \checkmark & \checkmark & 45.0 & 47.9 & 59.9 & 59.5 & 53.1 \\
        OnAVOS \cite{voigtlaender_OnAVOS_BMCV_2017} & \checkmark & \checkmark  & 46.6 & 51.4 & 60.1 & 62.7 & 55.2 \\
        DMM-Net\cite{zeng_dmmnet_iccv_2019} & \checkmark & \checkmark & 50.6 & 57.4 & 60.3 & 50.6 & 58.0 \\
        
        \midrule \midrule
        RGMP\cite{oh_seoung_fastvos_RGMP_CVPR_2018}& \checkmark &  & - & -  & 59.5 & 45.2 & 53.8 \\
        OSVOS \cite{caelles_OSVOS_CVPR_2017} & \checkmark &  & 54.2 & 60.7 & 59.8 & 60.5 & 58.8 \\
        GC\cite{li_gc_fast_eccv_2020}        & \checkmark &  & 68.9  & 75.7  & 72.6 & 75.6 & 73.2 \\
        AFB-URR\cite{liang_AFB_URR_NIPS_2020} & \checkmark &  & 74.1 & 82.6  & 78.8 & 83.1 & 79.6 \\
        GMVOS\cite{lu_episodicvos_eccv_2020} & \checkmark & & 74.0 & 80.9 & 80.7 & 85.1 & 80.2 \\
        LWLVOS$^{\ddagger}$\cite{bhat_learn_what_vos_ECCV_2020} & \checkmark &  & 76.4 & 84.4  & 80.4 & 84.9 & 81.5 \\
        KMN\cite{seong_kernelizedvos_eccv_2020} & \checkmark &  & 75.3  & 83.3 & 81.4 & 85.6 & 81.4 \\ 
        LCM\cite{hu_lcm_cvpr_2021}             & \checkmark &   & 75.7  & 83.4  & 82.2 & 86.7 & 82.0 \\
        \hline
        STM\cite{oh_STM_ICCV_2019} & \checkmark &  & 72.8 & 80.9  & 79.7 & 84.2 & 79.4 \\
        \textbf{Ours} & \checkmark &  & 75.1 & 83.4 & 79.0 & 83.5 & 80.3 \\

        \midrule \midrule
        A-GAME\cite{Johnander_AGame_CVPR_2019}&  &  & 60.8 & 66.2 & 67.8 & 69.5 & 66.1 \\
        FRTM\cite{Robinson_frtm-vos_CVPR_2020} &  &  & 76.2 & 74.1  & 72.3 & 76.2 & 72.1 \\
        LWLVOS$^{\ddagger}$\cite{bhat_learn_what_vos_ECCV_2020} & &  & 75.6 & 84.4  & 78.3 & 82.3 & 80.2 \\
        CFBI$^{\dagger}$ \cite{yang_CFBI_ECCV_2020}&  &  & 75.3 & 83.4 & 81.1 & 85.8 & 81.4 \\
        {RMNet\cite{xie_RMNet_CVPR_2021}}&  &  & {75.7} & {82.4} & {82.1} & {85.7} & {81.5} \\
        \hline
        STM\cite{oh_STM_ICCV_2019} &   &  & - & -  & - & - & 68.2 \\
        \textbf{Ours} &  &  & 70.6 & 77.9 & 78.5 & 82.7 & 77.4 \\

        \toprule
        \end{tabular}
    }
    \caption[Comparison for YouTube]{\textbf{Comparison on the YouTube-VOS18 validation set.}\footnotemark[1]}
    \label{table:result_youtube_18}
    \vspace{-\baselineskip}
\end{table}

\subsection{Ablation Study}
\label{sec::ablation_study}
In this section, we conduct several ablation experiments on the DAVIS17 validation set via training \texttt{protocol(2)} in \figref{fig:train_strategy_no_warp} to discuss the effectiveness of our proposed modules.
The baseline version of these experiments uses ResNet50, without any proposed modules.

\begin{figure}[!hb]
    \centering 
    \includegraphics[width=0.95\linewidth]{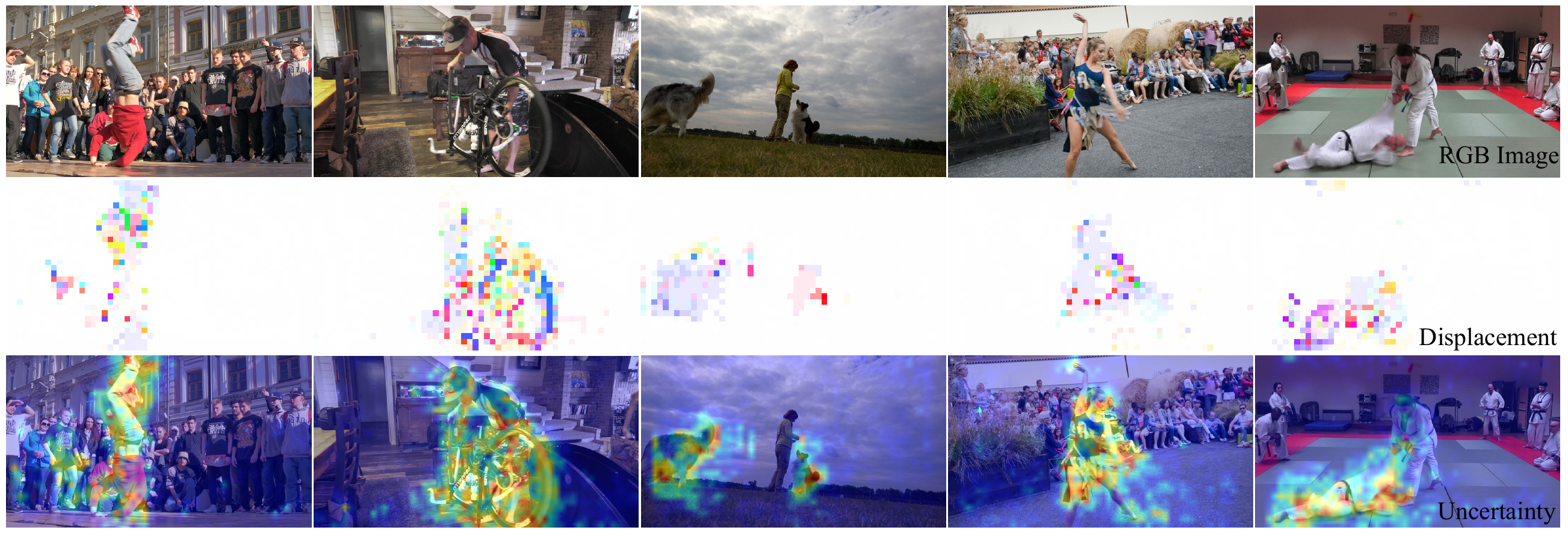}
    \vspace{-10pt}
    \caption{\textbf{Visualization of the displacement and uncertainty map}. We project the uncertainty map into a heat map and add it to the original image for better visualization.} %
    \label{fig:flow_uncertainty_shown}
\end{figure}

\begin{figure*}[!ht]
    \centering
    \includegraphics[width=0.97\linewidth]{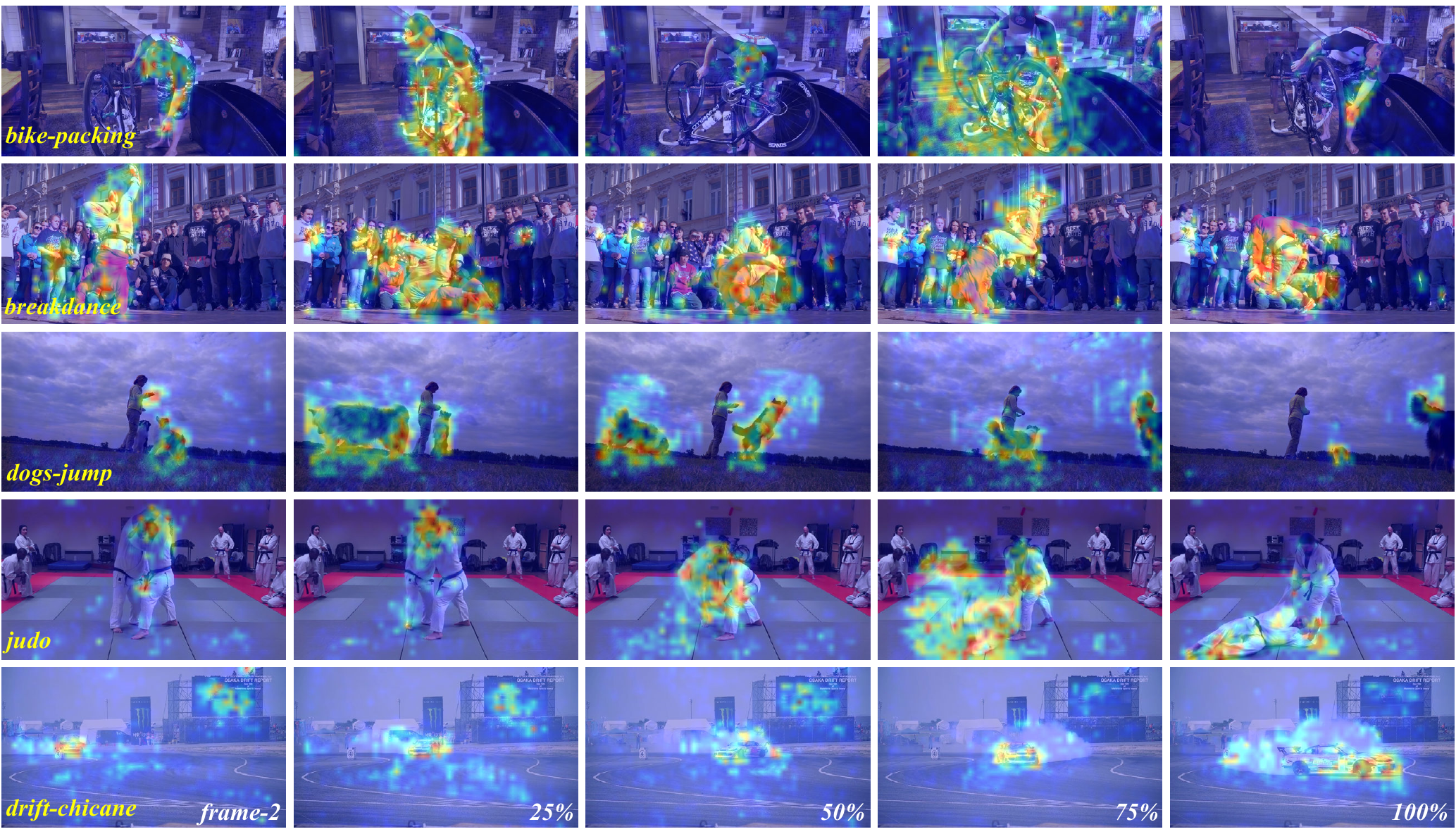}
    \vspace{-0.5\baselineskip}
    \caption{\textbf{Visualization of the uncertainty map of ours on the DAVIS17 validation set}. We project the uncertainty map into a heat map and add it to the original image for better visualization. We select one frame after every $1/4$ sequence length for visualization (frame-2 replace $0\%$). It can be verified that the moving regions in the adjacent images are activated with high uncertainty.}
    \label{fig:show_confidence_seqs}
    \vspace{-0.5\baselineskip}

\end{figure*}

\noindent\textbf{MU-Layer.}
We train a network without our MU-layer to study how it influences the matching based methods. As shown in \tabref{table:ablation_study} (\textit{a}) and (\textit{c}), our proposed MU-Layer can achieve $1.5\%$ \footnotesize($76.5\% \to 78.0\%$)\normalsize \  performance gain.
We also visualize the displacement and uncertainty map in \figref{fig:flow_uncertainty_shown} and \figref{fig:show_confidence_seqs}, which shows that our MU-Layer extracts reliable motion information.

\begin{figure}[!ht]
    \centering
    \includegraphics[width=\linewidth]{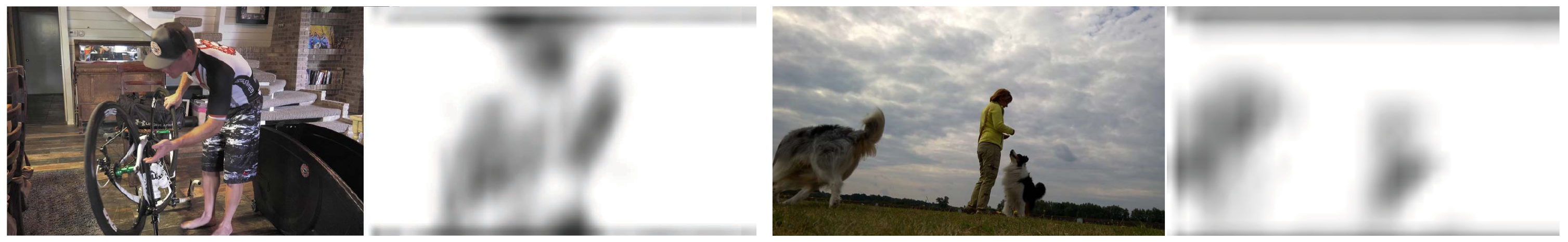}
    \vspace{-\baselineskip}
    \caption{\textbf{Visualization of the attention map}. We directly use the grayscale image to display the one-channel attention map.}  
    \label{fig:attention_map}
\end{figure}

\noindent\textbf{MASM.}
To evaluate the effectiveness of our fusion module, we report the result of addition (\textit{f}) and MASM (\textit{g}) in \tabref{table:ablation_study}, based on the complete structure (\textit{e}). 
The comparison shows that MASM improves the vanilla adding based fusion method by $1.0\%$ \footnotesize($80.1\% \to 81.1\%$)\normalsize \ in $\mathcal{J}$\&$\mathcal{F}$.
We visualize the spatial attention map of shape $H^{\prime} \times W^{\prime} \times 1$ before broadcast multiply and show it in \figref{fig:attention_map}. We can observe that the attention map focuses on the salient area of the moving objects, which demonstrates that MASM helps strengthen the object information.

\noindent\textbf{Segmentation Loss.}
We study how well-designed segmentation loss affects the performance. The results of (\textit{a}) and (\textit{b}) are listed in \tabref{table:ablation_study}, and the performance drops with or without such loss. This provides us a strong baseline and achieves better performance.
\begin{table}[!ht]
    \renewcommand\arraystretch{0.9}
    \setlength\tabcolsep{4pt}
    \centering
    \resizebox{0.78\linewidth}{!}{%
        \begin{tabular}{c|l|cccc}
        \toprule
         & Baseline and components & $\mathcal{J}$ & $\mathcal{F}$  & $\mathcal{J}$\&$\mathcal{F}$ & $\Delta$\\ 
        \toprule
        \rowcolor{lightgray} (\textit{a}) & w/o disp. (baseline)        & 74.5 & 78.4 & 76.5 & - \\
        (\textit{b}) & + BsCE\&IoU loss            & 76.2 & 80.9 & 78.6 & +2.1 \\
        (\textit{c}) & + disp. + uncertainty map   & 75.9 & 80.1 & 78.0 & +1.5 \\
        (\textit{d}) & + disp. + BsCE\&IoU         & 76.9 & 82.5 & 79.7 & +3.2 \\
        (\textit{e}) & + disp. + BsCE\&IoU + uncertainty & 78.3 & 83.9 & 81.1 & +4.6 \\
        \midrule \midrule
        (\textit{f}) & All three \textit{cmpt.} w [addition]   & 77.9 & 82.3 & 80.1 & +3.6 \\
        (\textit{g}) & All three \textit{cmpt.} w [motion attention]  & 78.3 & 83.9 & 81.1 & +4.6 \\
        \toprule
        \end{tabular}
    }
    \caption{\textbf{Ablation study of components.}  ``disp'': displacement, ``BsCE\&IoU'': using $\mathcal{L}_{bsce} \& \mathcal{L}_{miou}$ to supervise, ``[addition]'' and ``[motion attention]'' is the fusion method.} \label{table:ablation_study}
    \vspace{-0.5\baselineskip}
\end{table}

\subsection{Some Failure Cases}
\label{sec:failure_cases}
\begin{figure}[!ht]
    \centering
    \includegraphics[width=0.95\linewidth]{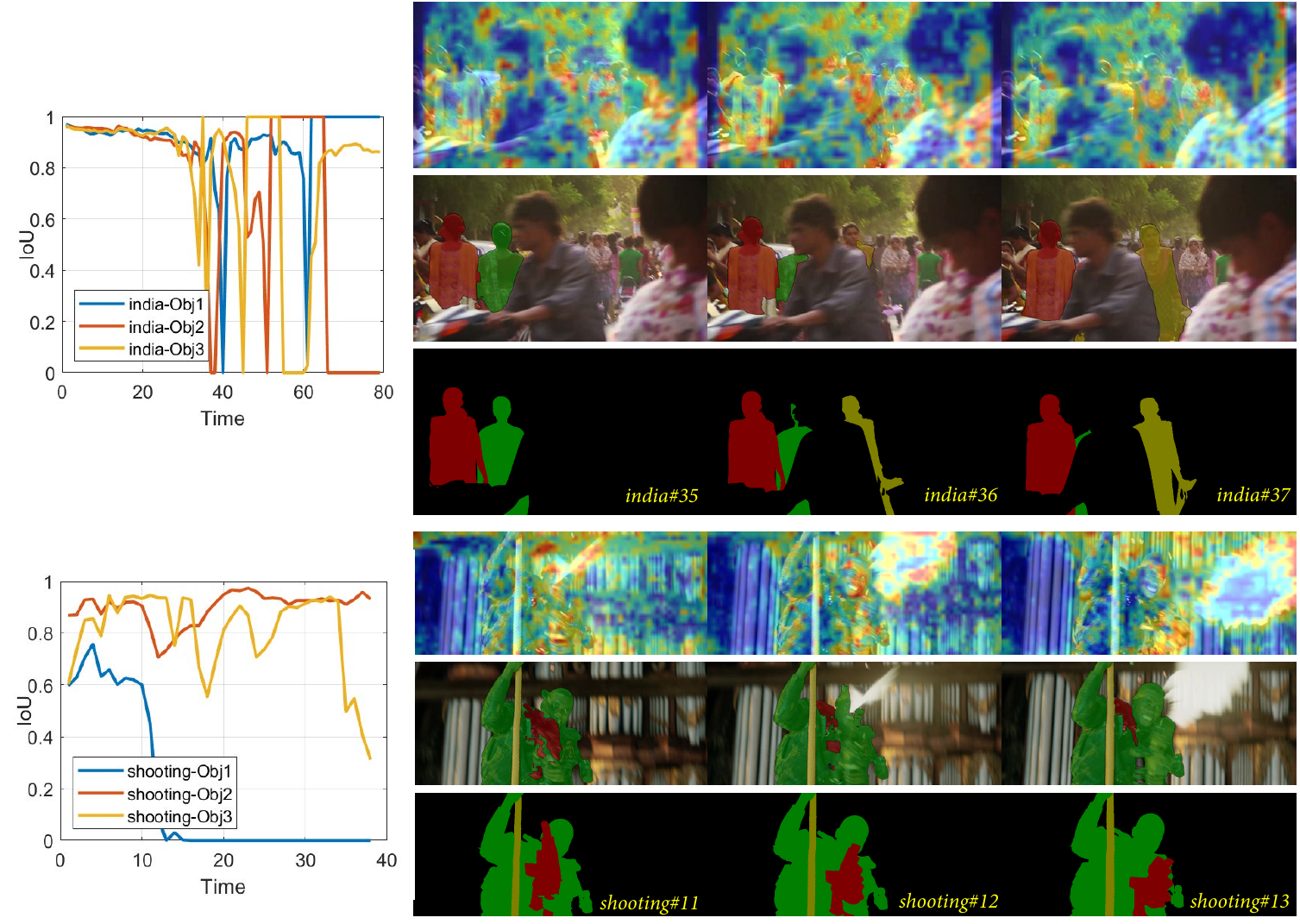}
    \vspace{-0.5\baselineskip}
    \caption{\textbf{Some failure cases of ours}. The left is the IoU per-frame over time. From top to bottom on the right are the uncertainty map, the predicted masks, and the ground truth.} \label{fig:failure_cases}
\end{figure}

Here we provide some failure cases of our scheme in \figref{fig:failure_cases}.
After the previous quantitative and qualitative comparisons, our solution has achieved great success by exploiting the effective motion information. 
As common optical flow networks cannot handle extremely complex/fast motions and large-scale occlusion well, and there is no ground truth for supervision in our implicit modeling method, we also have flaws in some extreme examples. However, our method still outperforms other competing methods under these challenging scenarios.
\vspace{-\baselineskip}

\begin{figure*}[!ht]
    \centering
    \includegraphics[width=0.98\linewidth]{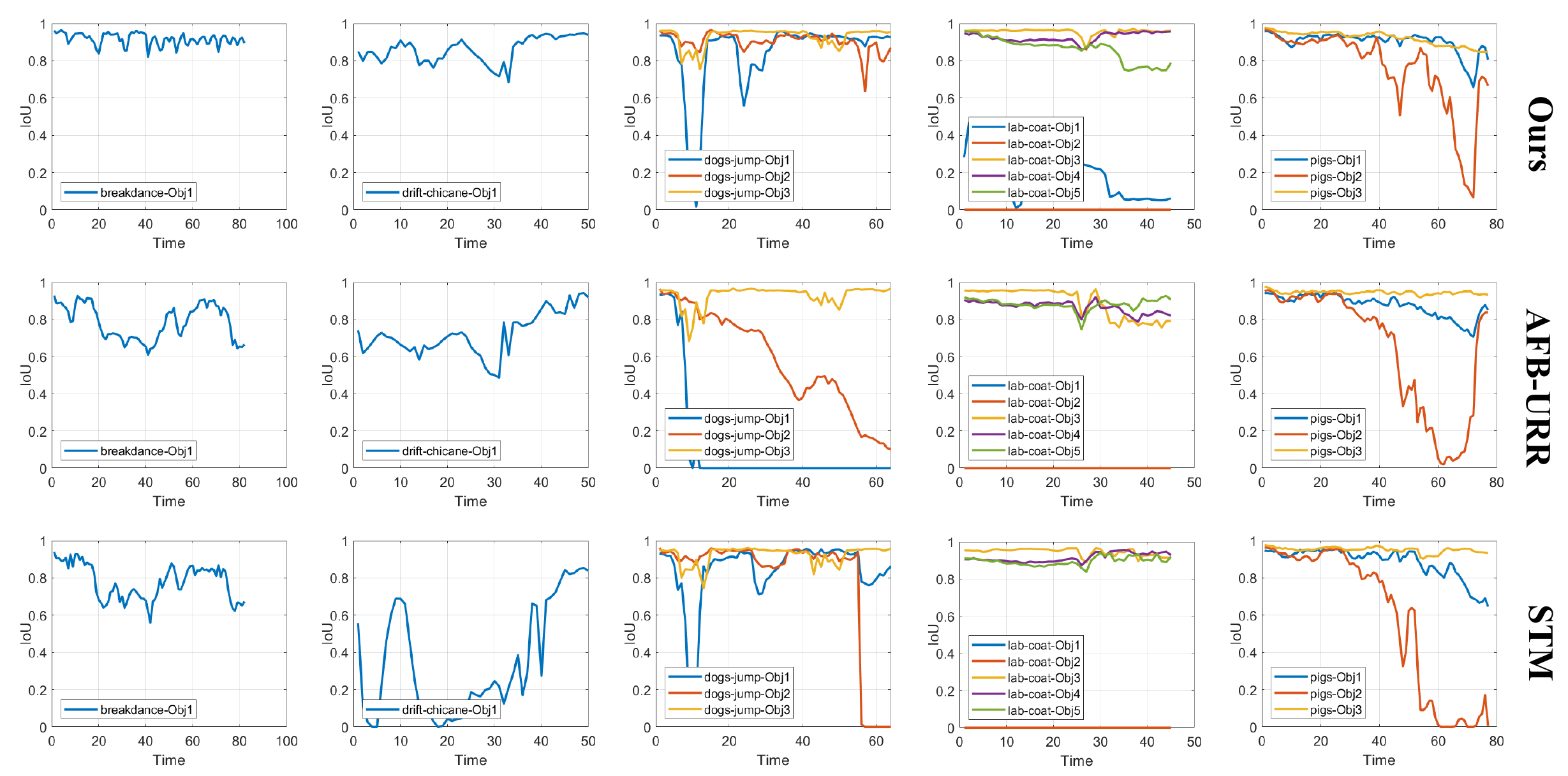}
    \vspace{-0.5\baselineskip}
    \caption{\textbf{IoU per-frame over time} of Ours, STM and AFB-URR on five video sequences from DAVIS17 validation set. The last three columns have multiple objects, while the first two columns have only a single object. Since we effectively use the motion information between adjacent frames, the IoU can remain high even in the latter part of the video sequences.}
    \label{fig:iou_vis}
\end{figure*}
\begin{figure*}[!ht]
\centering 
\includegraphics[width=0.98\linewidth]{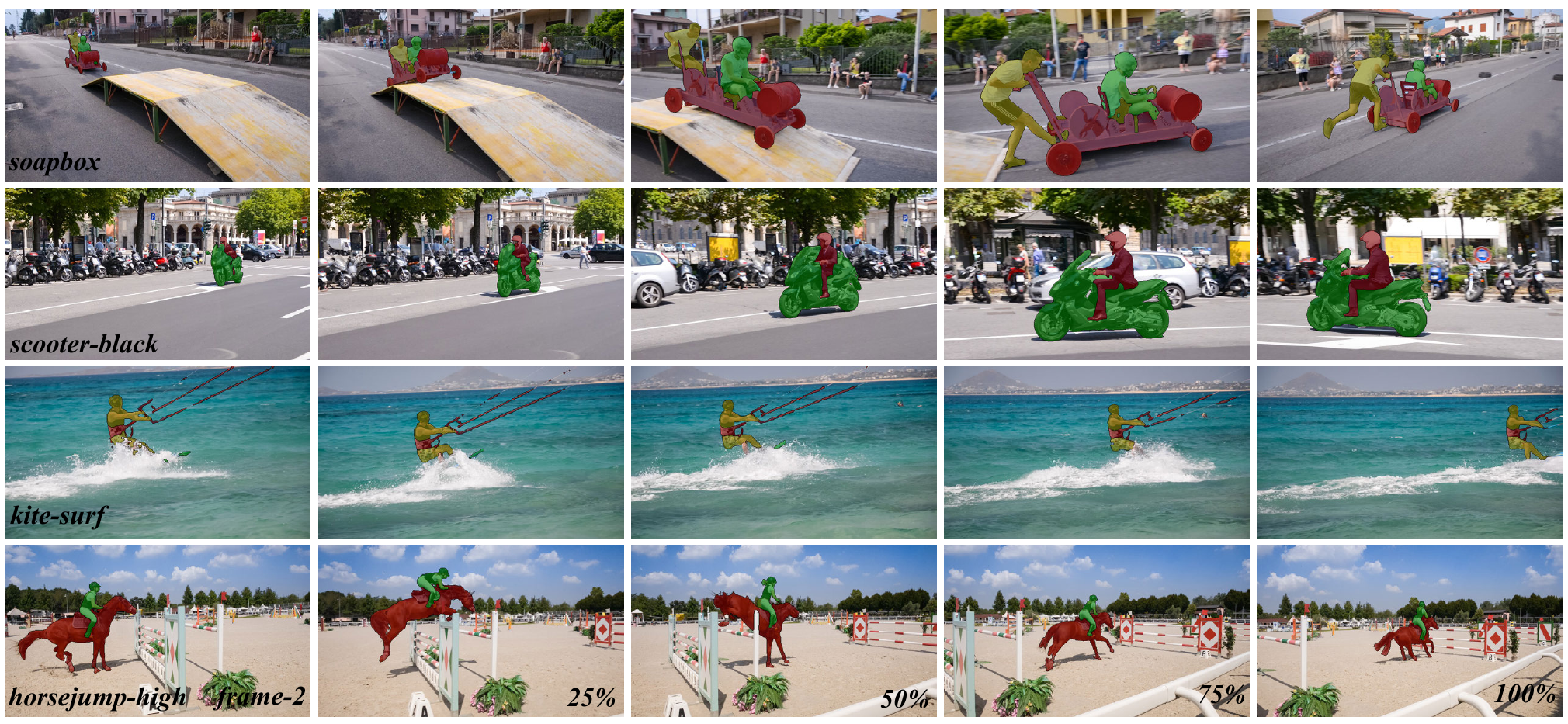}
\vspace{-0.5\baselineskip}
\caption{\textbf{Visualization of the quantitative results of ours on DAVIS17 validation set}. 
We select one frame every $1/4$ sequence length for visualization (frame-2 replace $0\%$). 
}
\label{fig:ours_quantitative_result}
\vspace{-0.5\baselineskip}

\end{figure*}
\begin{figure*}[!ht]
\centering
\includegraphics[width=\linewidth]{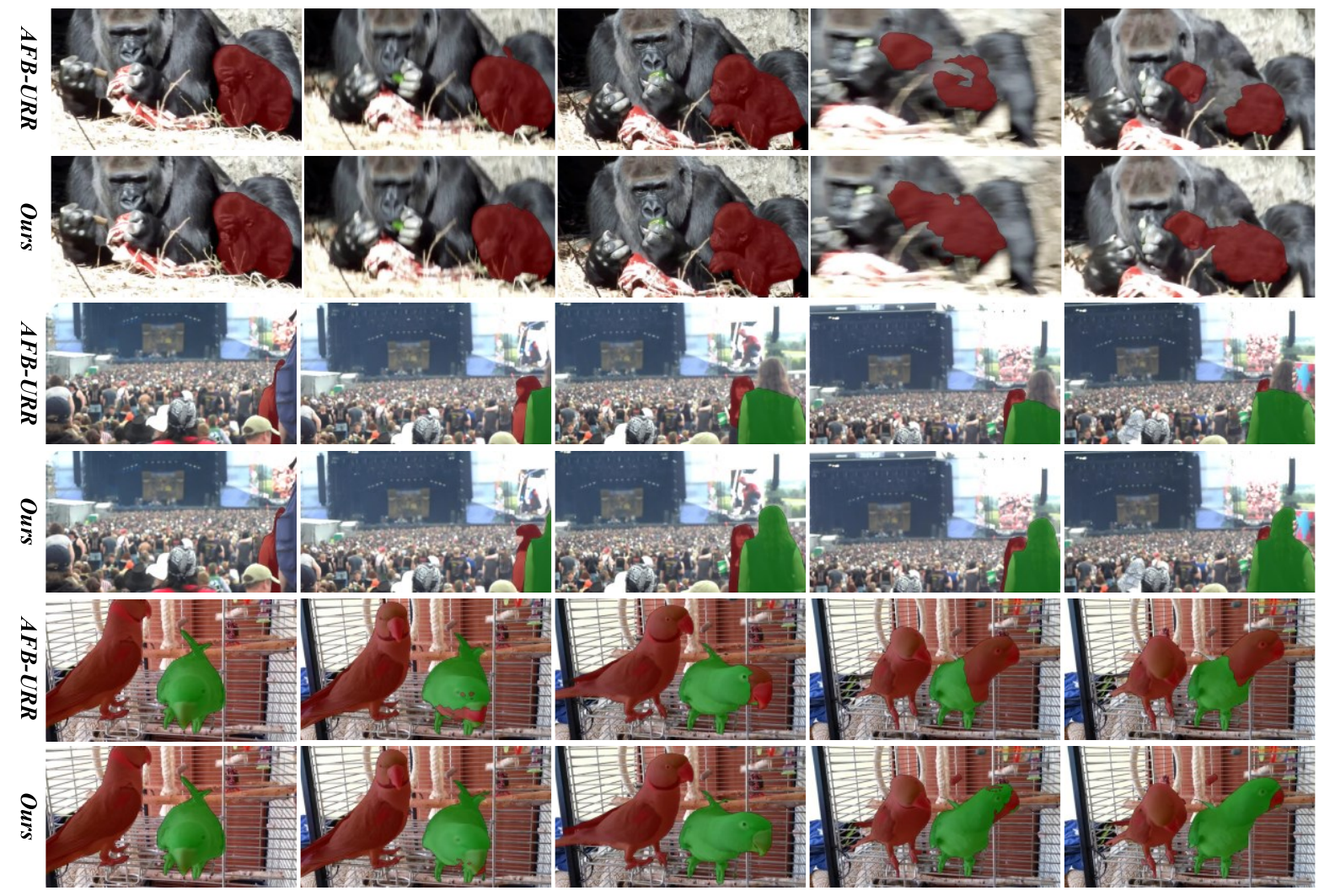}
\vspace{-1.5\baselineskip}
\caption{\textbf{Visualization of the quantitative results of ours and AFB-URR on the YouTube-VOS18 validation set}.}
\label{fig:YTB}
\end{figure*}

\subsection{Qualitative Comparison} \label{sec::qualitative_result}

We give a qualitative comparison with some \textit{SOTA} methods \cite{liang_AFB_URR_NIPS_2020,yang_CFBI_ECCV_2020,Robinson_frtm-vos_CVPR_2020,oh_STM_ICCV_2019} in \figref{fig:vis_compare_result}. In \textit{soapbox}\#74 and \textit{scooter-black}\#42, our method can handle the boundaries between objects well. In \textit{paragliding-launch}\#20, our method deals with moving thin lines better. 
However, FRTM\cite{Robinson_frtm-vos_CVPR_2020} lose parts of objects or over-cover the thin object.
AFB-URR\cite{liang_AFB_URR_NIPS_2020} struggles with discrimination between different objects and almost fails to detect small and thin objects.

In \figref{fig:iou_vis}, we show how IoU changes over time by per-frame evaluation. 
Because the motion information we used exists between any adjacent frames and will not disappear over time, our methods can still maintain high IoU in the last 50\% frames.
Compared with STM and AFB-URR, we have achieved more robust segmentation. 
More qualitative results of ours are shown in \figref{fig:ours_quantitative_result}. 
Thanks to the accurate modeling of motion information,
our method is robust to occlusion, complex/large motion of \textit{row\#1,2,4} and can distinguish the small/thin objects in \textit{row\#3}. 
And in \figref{fig:YTB}, we select 3 video sequences of YouTube-VOS18 for visual comparison with AFB-URR\cite{liang_AFB_URR_NIPS_2020}.

\section{Conclusion}
In this work, we advocated the return of motion information in the state-of-the-art dense-matching based semi-supervised VOS approaches. We proposed a novel motion uncertainty-aware pipeline for semi-supervised VOS, where the motion information is implicitly modeled. We implicitly built correlation volume by matching pixel-pairs between the reference frame and the query frame, enabling the learning of motion features. To address the challenging cases of occlusion and textureless regions, we incorporated the motion uncertainty into building dense correspondences. Furthermore, we proposed a motion-enhanced module to effectively fuse the motion feature and the semantic feature. Extensive experiments on benchmark datasets proved the superiority of our proposed framework.

\section*{Acknowledgements} This research was supported in part by the National Key Research and Development Program of China under Grant 2018AAA0102803 and the National Natural Science Foundation of China under Grants 61871325, 62001394 and 61901387.

\clearpage
\bibliography{7_References}

\begin{thebibliography}{10}
\expandafter\ifx\csname url\endcsname\relax
  \def\url#1{\texttt{#1}}\fi
\expandafter\ifx\csname urlprefix\endcsname\relax\def\urlprefix{URL }\fi
\expandafter\ifx\csname href\endcsname\relax
  \def\href#1#2{#2} \def\path#1{#1}\fi

\bibitem{liang_AFB_URR_NIPS_2020}
Y.~Liang, X.~Li, N.~Jafari, Q.~Chen, Video object segmentation with adaptive
  feature bank and uncertain-region refinement, in: Proceedings of the Advances
  in Neural Information Processing Systems (NeurIPS), 2020.

\bibitem{yang_CFBI_ECCV_2020}
Z.~Yang, Y.~Wei, Y.~Yang, Collaborative video object segmentation by
  foreground-background integration, in: Proceedings of the European Conference
  on Computer Vision (ECCV), 2020.

\bibitem{Robinson_frtm-vos_CVPR_2020}
A.~Robinson, F.~J. Lawin, M.~Danelljan, F.~S. Khan, M.~Felsberg, Learning fast
  and robust target models for video object segmentation, in: Proceedings of
  the IEEE Conference on Computer Vision and Pattern Recognition (CVPR), 2020.

\bibitem{oh_STM_ICCV_2019}
S.~W. Oh, J.-Y. Lee, N.~Xu, S.~J. Kim, Video object segmentation using
  space-time memory networks, in: Proceedings of the IEEE International
  Conference on Computer Vision (ICCV), 2019, pp. 9226--9235.

\bibitem{yin2021agunet_PR}
Y.~Yin, D.~Xu, X.~Wang, L.~Zhang, Agunet: Annotation-guided u-net for fast
  one-shot video object segmentation, Pattern Recognition 110 (2021) 107580.

\bibitem{sun2020adaptive_PR}
M.~Sun, J.~Xiao, E.~G. Lim, Y.~Xie, J.~Feng, Adaptive roi generation for video
  object segmentation using reinforcement learning, Pattern Recognition 106
  (2020) 107465.

\bibitem{zhao2021real_PR}
Z.~Zhao, S.~Zhao, J.~Shen, Real-time and light-weighted unsupervised video
  object segmentation network, Pattern Recognition (2021) 108120.

\bibitem{Perazzi_masktrack_CVPR_2017}
F.~Perazzi, A.~Khoreva, R.~Benenson, B.~Schiele, A.Sorkine-Hornung, Learning
  video object segmentation from static images, in: Proceedings of the IEEE
  Conference on Computer Vision and Pattern Recognition (CVPR), 2017.

\bibitem{oh_seoung_fastvos_RGMP_CVPR_2018}
S.~W. Oh, J.-Y. Lee, K.~Sunkavalli, S.~J. Kim, Fast video object segmentation
  by reference-guided mask propagation, in: Proceedings of the IEEE Conference
  on Computer Vision and Pattern Recognition (CVPR), 2018, pp. 7376--7385.

\bibitem{li_vsreid_cvprw_davis_2017}
X.~Li, Y.~Qi, Z.~Wang, K.~Chen, Z.~Liu, J.~Shi, P.~Luo, X.~Tang, C.~C. Loy,
  Video object segmentation with re-identification, in: The 2017 DAVIS
  Challenge on Video Object Segmentation - CVPR Workshops, 2017.

\bibitem{Li_STM-cycle_NeurIPS_2020}
Y.~Li, N.~Xu, P.~Jinlong, J.~See, L.~Weiyao, Delving into the cyclic mechanism
  in semi-supervised video object segmentation, in: Proceedings of the Advances
  in Neural Information Processing Systems (NeurIPS), 2020.

\bibitem{lu_episodicvos_eccv_2020}
X.~Lu, W.~Wang, D.~Martin, T.~Zhou, J.~Shen, V.~G. Luc, Video object
  segmentation with episodic graph memory networks, in: Proceedings of the
  European Conference on Computer Vision (ECCV), 2020.

\bibitem{seong_kernelizedvos_eccv_2020}
H.~Seong, J.~Hyun, E.~Kim, Kernelized memory network for video object
  segmentation, in: Proceedings of the European Conference on Computer Vision
  (ECCV), 2020, pp. 629--645.

\bibitem{liyu_fast_vos_ECCV_2020}
Y.~Li, Z.~Shen, Y.~Shan, Fast video object segmentation using the global
  context module, in: Proceedings of the European Conference on Computer Vision
  (ECCV), Springer, 2020, pp. 735--750.

\bibitem{cheng_segflow_cvpr_2017}
J.~Cheng, Y.-H. Tsai, S.~Wang, M.-H. Yang, Segflow: Joint learning for video
  object segmentation and optical flow, in: Proceedings of the IEEE
  International Conference on Computer Vision (ICCV), 2017, pp. 686--695.

\bibitem{jang_online_vos_ctn_cvpr_2017}
W.-D. Jang, C.-S. Kim, Online video object segmentation via convolutional
  trident network, in: Proceedings of the IEEE Conference on Computer Vision
  and Pattern Recognition (CVPR), 2017, pp. 5849--5858.

\bibitem{Dave_seg_anything_moves_ICCV_2019}
A.~Dave, P.~Tokmakov, D.~Ramanan, Towards segmenting anything that moves, in:
  Proceedings of the IEEE International Conference on Computer Vision Workshops
  (ICCVW), 2019.

\bibitem{khoreva_lucid_dream_DAVIS_2017}
A.~Khoreva, R.~Benenson, E.~Ilg, T.~Brox, B.~Schiele, Lucid data dreaming for
  object tracking, in: The 2017 DAVIS Challenge on Video Object Segmentation -
  CVPR Workshops, 2017.

\bibitem{tian2020joint_PR}
Y.~Tian, G.~Cheng, J.~Gelernter, S.~Yu, C.~Song, B.~Yang, Joint temporal
  context exploitation and active learning for video segmentation, Pattern
  Recognition 100 (2020) 107158.

\bibitem{ilg_flownet2_cvpr_2017}
E.~Ilg, N.~Mayer, T.~Saikia, M.~Keuper, A.~Dosovitskiy, T.~Brox, Flownet 2.0:
  Evolution of optical flow estimation with deep networks, in: Proceedings of
  the IEEE Conference on Computer Vision and Pattern Recognition (CVPR), 2017,
  pp. 2462--2470.

\bibitem{hui_liteflownet_cvpr_2018}
T.-W. Hui, X.~Tang, C.~C. Loy, Liteflownet: A lightweight convolutional neural
  network for optical flow estimation, in: Proceedings of the IEEE Conference
  on Computer Vision and Pattern Recognition (CVPR), 2018, pp. 8981--8989.

\bibitem{revaud_epicflow_cvpr_2015}
J.~Revaud, P.~Weinzaepfel, Z.~Harchaoui, C.~Schmid, Epicflow: Edge-preserving
  interpolation of correspondences for optical flow, in: Proceedings of the
  IEEE Conference on Computer Vision and Pattern Recognition (CVPR), 2015, pp.
  1164--1172.

\bibitem{hu_efficient_flow_cvpr_2016}
Y.~Hu, R.~Song, Y.~Li, Efficient coarse-to-fine patchmatch for large
  displacement optical flow, in: Proceedings of the IEEE Conference on Computer
  Vision and Pattern Recognition (CVPR), 2016, pp. 5704--5712.

\bibitem{kroeger_fast_flow_eccv_2016}
T.~Kroeger, R.~Timofte, D.~Dai, L.~Van~Gool, Fast optical flow using dense
  inverse search, in: Proceedings of the European Conference on Computer Vision
  (ECCV), Springer, 2016, pp. 471--488.

\bibitem{wang_DICL_NIPS_2020}
J.~Wang, Y.~Zhong, Y.~Dai, K.~Zhang, P.~Ji, H.~Li, Displacement-invariant
  matching cost learning for accurate optical flow estimation, arXiv preprint
  arXiv:2010.14851 (2020).

\bibitem{teed_raft_eccv_2020}
Z.~Teed, J.~Deng, Raft: Recurrent all-pairs field transforms for optical flow,
  in: Proceedings of the European Conference on Computer Vision (ECCV),
  Springer, 2020, pp. 402--419.

\bibitem{Ziqin_RANet_CVPR_2019}
Z.~Wang, J.~Xu, L.~Liu, F.~Zhu, L.~Shao, Ranet: Ranking attention network for
  fast video object segmentation, in: Proceedings of the IEEE Conference on
  Computer Vision and Pattern Recognition (CVPR), 2019, pp. 3977--3986.

\bibitem{xie_RMNet_CVPR_2021}
H.~Xie, H.~Yao, S.~Zhou, S.~Zhang, W.~Sun, Efficient regional memory network
  for video object segmentation, in: Proceedings of the IEEE Conference on
  Computer Vision and Pattern Recognition (CVPR), 2021, pp. 1286--1295.

\bibitem{kaiming_resnet50_cvpr_2016}
K.~He, X.~Zhang, S.~Ren, J.~Sun, Deep residual learning for image recognition,
  in: Proceedings of the IEEE Conference on Computer Vision and Pattern
  Recognition (CVPR), 2016, pp. 770--778.

\bibitem{li_MotionAttention_ICCV_2019}
H.~Li, G.~Chen, G.~Li, Y.~Yu, Motion guided attention for video salient object
  detection, in: Proceedings of the IEEE International Conference on Computer
  Vision (ICCV), 2019, pp. 7274--7283.

\bibitem{zhou_matnet_aaai_2020_tip}
T.~Zhou, J.~Li, S.~Wang, R.~Tao, J.~Shen, Matnet: Motion-attentive transition
  network for zero-shot video object segmentation, IEEE Transactions on Image
  Processing (TIP) 29 (2020) 8326--8338.

\bibitem{wu_bridging_bootstrap_celoss_2016}
Z.~Wu, C.~Shen, A.~v.~d. Hengel, Bridging category-level and instance-level
  semantic image segmentation, arXiv preprint arXiv:1605.06885 (2016).

\bibitem{paszke_pytorch_NIPS_2019}
A.~Paszke, S.~Gross, F.~Massa, A.~Lerer, J.~Bradbury, G.~Chanan, T.~Killeen,
  Z.~Lin, N.~Gimelshein, L.~Antiga, et~al., Pytorch: An imperative style,
  high-performance deep learning library, in: Proceedings of the Advances in
  Neural Information Processing Systems (NeurIPS), 2019, pp. 8026--8037.

\bibitem{loshchilov_AdamW_ICLR_2018}
I.~Loshchilov, F.~Hutter, Decoupled weight decay regularization, in:
  Proceedings of the The International Conference on Learning Representations
  (ICLR), 2019.

\bibitem{Pont_DAVIS_arxiv_2017}
J.~Pont-Tuset, F.~Perazzi, S.~Caelles, P.~Arbel{\'a}ez, A.~Sorkine-Hornung,
  L.~Van~Gool, The 2017 davis challenge on video object segmentation, arXiv
  preprint arXiv:1704.00675 (2017).

\bibitem{Xu_YouTube_ECCV_2018}
N.~Xu, L.~Yang, Y.~Fan, J.~Yang, D.~Yue, Y.~Liang, B.~Price, S.~Cohen,
  T.~Huang, Youtube-vos: Sequence-to-sequence video object segmentation, in:
  Proceedings of the European Conference on Computer Vision (ECCV), 2018, pp.
  585--601.

\bibitem{Cheng_PAMI_2015}
M.-M. Cheng, N.~J. Mitra, X.~Huang, P.~H.~S. Torr, S.-M. Hu, Global contrast
  based salient region detection, IEEE Transactions on Pattern Analysis and
  Machine Intelligence (T-PAMI) 37~(3) (2015) 569--582.

\bibitem{everingham_pascal_voc_2010}
M.~Everingham, L.~Van~Gool, C.~K. Williams, J.~Winn, A.~Zisserman, The pascal
  visual object classes (voc) challenge, International Journal of Computer
  Vision (IJCV) 88~(2) (2010) 303--338.

\bibitem{shi_hierarchical_pami_2015}
J.~Shi, Q.~Yan, L.~Xu, J.~Jia, Hierarchical image saliency detection on
  extended cssd, IEEE Transactions on Pattern Analysis and Machine Intelligence
  (T-PAMI) 38~(4) (2015) 717--729.

\bibitem{lin_coco_microsoft_2014}
T.-Y. Lin, M.~Maire, S.~Belongie, J.~Hays, P.~Perona, D.~Ramanan,
  P.~Doll{\'a}r, C.~L. Zitnick, Microsoft coco: Common objects in context, in:
  Proceedings of the European Conference on Computer Vision (ECCV), Springer,
  2014, pp. 740--755.

\bibitem{li_secrets_2014}
Y.~Li, X.~Hou, C.~Koch, J.~M. Rehg, A.~L. Yuille, The secrets of salient object
  segmentation, in: Proceedings of the IEEE Conference on Computer Vision and
  Pattern Recognition (CVPR), 2014, pp. 280--287.

\bibitem{yang_osmn_cvpr_2018}
L.~Yang, Y.~Wang, X.~Xiong, J.~Yang, A.~K. Katsaggelos, Efficient video object
  segmentation via network modulation, in: Proceedings of the IEEE Conference
  on Computer Vision and Pattern Recognition (CVPR), 2018, pp. 6499--6507.

\bibitem{voigtlaender_OnAVOS_BMCV_2017}
P.~Voigtlaender, B.~Leibe, Online adaptation of convolutional neural networks
  for video object segmentation, in: Proceedings of the British Machine Vision
  Conference (BMVC), 2017.

\bibitem{voigtlaender_feelvos_CVPR_2019}
P.~Voigtlaender, Y.~Chai, F.~Schroff, H.~Adam, B.~Leibe, L.-C. Chen, Feelvos:
  Fast end-to-end embedding learning for video object segmentation, in:
  Proceedings of the IEEE Conference on Computer Vision and Pattern Recognition
  (CVPR), 2019, pp. 9481--9490.

\bibitem{bhat_learn_what_vos_ECCV_2020}
G.~Bhat, F.~J. Lawin, M.~Danelljan, A.~Robinson, M.~Felsberg, L.~Van~Gool,
  R.~Timofte, Learning what to learn for video object segmentation, in:
  Proceedings of the European Conference on Computer Vision (ECCV), 2020, pp.
  777--794.

\bibitem{Johnander_AGame_CVPR_2019}
J.~Johnander, M.~Danelljan, E.~Brissman, F.~S. Khan, M.~Felsberg, A generative
  appearance model for end-to-end video object segmentation, in: Proceedings of
  the IEEE Conference on Computer Vision and Pattern Recognition (CVPR), 2019.

\bibitem{cho_pmvos_arxiv_2020}
S.~Cho, H.~Lee, S.~Woo, S.~Jang, S.~Lee, Pmvos: Pixel-level matching-based
  video object segmentation, arXiv preprint arXiv:2009.08855 (2020).

\bibitem{maninis_OSVOS-S_CVPR_2018}
K.-K. Maninis, S.~Caelles, Y.~Chen, J.~Pont-Tuset, L.~Leal-Taix{\'e},
  D.~Cremers, L.~Van~Gool, Video object segmentation without temporal
  information, IEEE Transactions on Pattern Analysis and Machine Intelligence
  (T-PAMI) 41~(6) (2018) 1515--1530.

\bibitem{bao_cinm_cvpr_2018}
L.~Bao, B.~Wu, W.~Liu, Cnn in mrf: Video object segmentation via inference in a
  cnn-based higher-order spatio-temporal mrf, in: Proceedings of the IEEE
  Conference on Computer Vision and Pattern Recognition (CVPR), 2018, pp.
  5977--5986.

\bibitem{li_gc_fast_eccv_2020}
Y.~Li, Z.~Shen, Y.~Shan, Fast video object segmentation using the global
  context module, in: Proceedings of the European Conference on Computer Vision
  (ECCV), Springer, 2020, pp. 735--750.

\bibitem{hu_lcm_cvpr_2021}
L.~Hu, P.~Zhang, B.~Zhang, P.~Pan, Y.~Xu, R.~Jin, Learning position and target
  consistency for memory-based video object segmentation, in: Proceedings of
  the IEEE Conference on Computer Vision and Pattern Recognition (CVPR), 2021,
  pp. 4144--4154.

\bibitem{luiten_premvos_accv_2018}
J.~Luiten, P.~Voigtlaender, B.~Leibe, Premvos: Proposal-generation, refinement
  and merging for video object segmentation, in: Proceedings of the Asian
  Conference on Computer Vision (ACCV), Springer, 2018, pp. 565--580.

\bibitem{zeng_dmmnet_iccv_2019}
X.~Zeng, R.~Liao, L.~Gu, Y.~Xiong, S.~Fidler, R.~Urtasun, Dmm-net:
  Differentiable mask-matching network for video object segmentation, in:
  Proceedings of the IEEE International Conference on Computer Vision (ICCV),
  2019, pp. 3929--3938.

\bibitem{caelles_OSVOS_CVPR_2017}
S.~Caelles, K.-K. Maninis, J.~Pont-Tuset, L.~Leal-Taix{\'e}, D.~Cremers,
  L.~Van~Gool, One-shot video object segmentation, in: Proceedings of the IEEE
  Conference on Computer Vision and Pattern Recognition (CVPR), 2017, pp.
  221--230.

\end{thebibliography}

\end{document}